\documentclass[10pt,twocolumn,letterpaper]{article}

\usepackage[pagenumbers]{cvpr} 

\usepackage{graphicx}
\usepackage{amsmath}
\usepackage{amssymb}
\usepackage{booktabs}

\usepackage[pagebackref,breaklinks,colorlinks]{hyperref}

\usepackage[capitalize]{cleveref}
\crefname{section}{Sec.}{Secs.}
\Crefname{section}{Section}{Sections}
\Crefname{table}{Table}{Tables}
\crefname{table}{Tab.}{Tabs.}

\def\cvprPaperID{9032} 
\def\confName{CVPR}
\def\confYear{2022}

\usepackage[utf8]{inputenc} 
\usepackage[T1]{fontenc}    
\usepackage{hyperref}       
\usepackage{url}            
\usepackage{booktabs}       
\usepackage{amsfonts}       
\usepackage{nicefrac}       
\usepackage{microtype}      

\usepackage{microtype}
\usepackage{wrapfig}
\usepackage{graphicx}
\usepackage{booktabs} 
\usepackage{times}
\usepackage{epsfig}
\usepackage{graphicx}
\usepackage{amsmath}
\usepackage{amssymb}
\usepackage{amsthm}
\usepackage{amsfonts}
\usepackage{bbm}
\usepackage{multirow}
\usepackage{booktabs}
\usepackage{threeparttable}

\usepackage[dvipsnames]{xcolor}
\colorlet{darkgreen}{green!65!black}
\colorlet{darkblue}{blue!75!black}
\colorlet{darkred}{red!80!black}
\definecolor{lightblue}{HTML}{0071bc}
\definecolor{lightgreen}{HTML}{39b54a}

\newtheorem{lemma}{Lemma}

\newtheorem{definition}{Definition}
\newtheorem{corollary}{Corollary}

\newcommand{\name} {PCL}

\newcommand{\red}[1]{\textcolor{black}{#1}}
\newcommand{\blue}[1]{\textcolor{black}{#1}}

\newenvironment{Itemize}%
{
\setlength{\leftmargini}{9pt}%
\begin{itemize}%
\setlength{\itemsep}{0pt}%
\setlength{\topsep}{0pt}%
\setlength{\partopsep}{0pt}%
\setlength{\parskip}{0pt}}%
{\end{itemize}}

{
\setlength{\leftmargini}{9pt}%
\begin{enumerate}%
\setlength{\itemsep}{0pt}%
\setlength{\topsep}{0pt}%
\setlength{\partopsep}{0pt}%
\setlength{\parskip}{0pt}}%
{\end{enumerate}}

\graphicspath{{./figures/}}

\begin{document}


\title{Addressing Feature Suppression in Unsupervised Visual Representations}

\author{Tianhong Li$^{1,}$\thanks{Indicates equal contribution.} \quad Lijie Fan$^{1,{\ast}}$ \quad Yuan Yuan$^1$ \quad Hao He$^1$ \quad Yonglong Tian$^1$ \\ 
\quad Rogerio Feris$^2$ \quad Piotr Indyk$^1$ \quad Dina Katabi$^1$ \\\\ $^1$MIT CSAIL, $^2$MIT-IBM Watson AI Lab}

\maketitle

 \begin{abstract}
 Contrastive learning is one of the fastest growing research areas in machine learning due to its ability to learn useful representations without labeled data. However, contrastive learning is susceptible to \red{feature suppression} – i.e., it may discard important information relevant to the task of interest, and learn irrelevant features. Past work has addressed this limitation via handcrafted data augmentations that eliminate irrelevant information. This approach however does not work across all datasets and tasks. Further, data augmentations fail in addressing \red{feature suppression} in multi-attribute classification when one attribute \red{can suppress features relevant to} other attributes.  In this paper, we analyze the objective function of contrastive learning and formally prove that it is vulnerable to \red{feature suppression}. We then present predictive contrastive learning (\name), a framework for learning unsupervised representations that are robust to \red{feature suppression}. The key idea is to force the learned representation to predict the input, and hence prevent it from discarding important information. Extensive experiments verify that \name\ is robust to \red{feature suppression} and outperforms state-of-the-art contrastive learning methods on a variety of datasets and tasks.

 \end{abstract}

 \section{Introduction}
The area of unsupervised or self-supervised representation learning is growing rapidly~\cite{he2020momentum,doersch2015unsupervised,ye2019unsupervised,hjelm2018learning,grill2020bootstrap,bachman2019learning,zhuang2019local,misra2020self,han2020memory}.
It refers to learning data representations that capture potential labels of interest, and doing so without human supervision. Contrastive learning is increasingly considered as a standard and highly competitive method for unsupervised representation learning. Features learned with this method have been shown to generalize well to downstream tasks,  and in some cases surpass the performance of supervised models~\cite{oord2018representation,caron2020unsupervised,tschannen2019mutual,chen2020simple,chen2020big,chen2020improved}. 
 
Contrastive learning learns representations by contrasting positive samples against negative samples. During training, a data sample is chosen as an anchor (e.g., an image);  positive samples are chosen as different augmented versions of the anchor (e.g., randomly cropping and color distorting the image), whereas negative samples come from other samples in the dataset. 
 
Yet contrastive learning is vulnerable to \red{feature suppression} -- i.e., if simple features are contrastive enough to separate positive samples from negative samples, contrastive learning might learn such simple (or simpler) features even if irrelevant to the tasks of interest, and \red{other more relevant features are suppressed}. For example, the authors of \cite{chen2020simple} show that color distribution can be used to distinguish patches cropped from the same image, from patches from different images; yet such feature is not useful for object classification. Past work addresses this problem by designing handcrafted data augmentations that eliminate the irrelevant features, so that the network may learn the relevant information \cite{he2020momentum,chen2020simple,chen2020big,chen2020improved,chen2020intriguing}. 

\begin{figure}
\begin{center}
\begin{tabular}{cc}
\label{fig:intro_rf}
\includegraphics[width=0.2\textwidth]{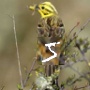} & \hspace*{0.13in}
\includegraphics[width=0.2\textwidth]{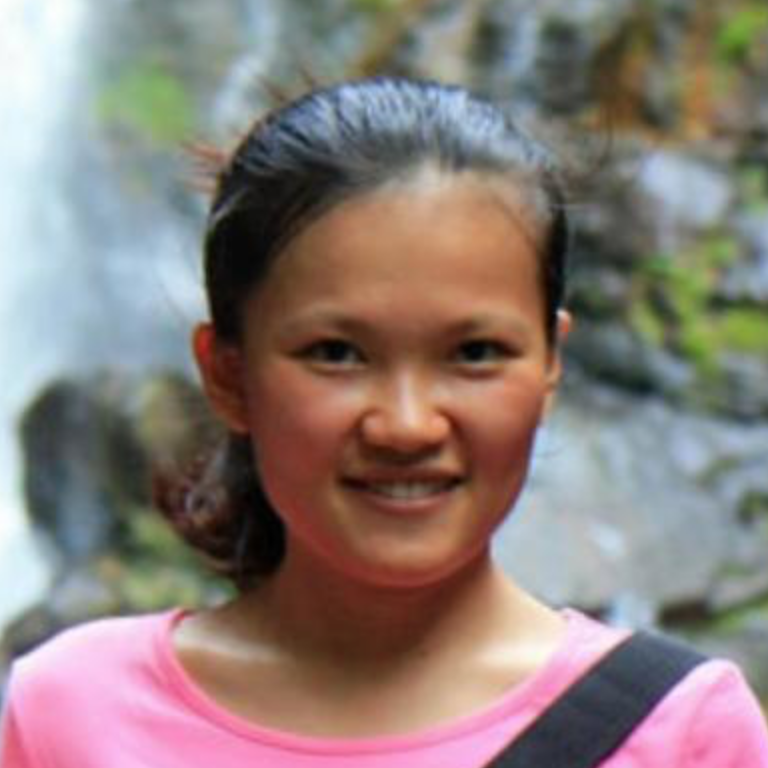}\\
{\footnotesize (a) Digit \& Bkgd} & \hspace*{0.13in} {\footnotesize (b) Face Attribute}
\end{tabular}
\end{center}
\vspace{-10pt}
\caption{\footnotesize{
(a) In Colorful-Moving-MNIST \cite{tian2020makes}, the input has two types of information: digit and background object. But contrastive learning methods focus on the background object and ignore the digit. (b) Each image in FairFace \cite{karkkainen2019fairface} has multiple attributes such as age, gender, ethnicity, etc. Existing contrastive learning methods focus on ethnicity and partially ignore other attributes.}}
\label{fig:intro}
\vspace{-10pt}
\end{figure}

However, in many scenarios it is hard to design augmentations to solve \red{the problem of feature suppression}. For example, the authors of \cite{tian2020makes} highlight the scenario in Figure~\ref{fig:intro} (a), where each image shows a digit (from MNIST) on a randomly chosen background object (from STL-10).  They show that features related to background objects can create a shortcut that \red{prevent contrastive learning from learning features related to digits}. In this case, one cannot simply eliminate the background information since such a design, though would help digit classification, would harm the background classification task. %
A similar problem exists in the task of human face attribute classification, where each face image can be used in multiple downstream tasks including gender, age, and ethnicity classification (Figure~\ref{fig:intro} (b)), but the features learned by contrastive learning can be biased to only one of the attributes (e.g., ethnicity) and show poor performance on other attributes (gender and age) as shown in the experiments section. It is hard to come up with data augmentations that eliminate the dominant attribute without harming the corresponding classification task. Moreover, as machine learning keeps expanding to new modalities it becomes increasingly difficult to design handcrafted data augmentations because many new modalities are hard to directly interpret by humans (e.g., acceleration from wearable devices), or the interpretation requires domain experts (e.g., medical data).

\red{In this paper, we first provide a theoretical analysis of contrastive learning and prove it is vulnerable to feature suppression. Our analysis shows that even with large feature dimensions, contrastive learning has many local minimums that discard significant information about the input, and hence cause feature suppression. Furthermore, the value of the loss function at such local minimums is very close to its value at the global minimum, making it hard to propel the model out of such local minimums. } 

\red{Second, we propose predictive contrastive learning (\name) as a training scheme that prevents feature suppression. \name\ learns representations using contrastive and \red{predictive} learning simultaneously. We use the term predictive learning to refer to tasks that force the representation to predict the input, such as inpainting, colorization, or autoencoding. Such tasks counter the effect of feature suppression because they force the learned features to retain the information in the input. More formally, if the contrastive loss (i.e., the InfoNCE loss) gets stuck in a local minimum that loses semantic information, the predictive loss naturally becomes very high, forcing the model to exit such local minimums. An interesting feature of \name\ is that the predictive task is used only during training, and hence introduces no computation overhead during testing. }


We evaluate \name~and compare it with state-of-the-art contrastive learning baselines on four different datasets: ImageNet,  MPII \cite{andriluka20142d}, Colorful-Moving-MNIST \cite{tian2020makes}, and FairFace \cite{karkkainen2019fairface}. For all tasks, \name~achieves superior performance and outperforms the state-of-the-art baselines by large margins, demonstrating robustness against \red{feature suppression}.

The paper makes the following contributions:
\begin{Itemize}
	\item \blue{It provides a theoretical analysis of contrastive learning that proves its vulnerability to feature suppression.}
    \item It introduces \name, an \red{unsupervised} learning framework that automatically avoids \red{feature suppression} and provides a representation that learns all of the semantics in the input and can support different downstream tasks and multi-attribute classification.       
    \item It empirically shows that SOTA contrastive learning baselines (e.g., SimCLR, MoCo, and BYOL) suffer from \red{feature suppression}, and that \name\  outperforms those baselines on several important tasks including object recognition, pose estimation, and face attribute classification.
\end{Itemize}

 \section{Related Work}

Early work on unsupervised representation learning has focused on designing pretext tasks and training the network to predict their pseudo labels. Such tasks include solving jigsaw puzzles \cite{noroozi2016unsupervised}, restoring a missing patch in the input \cite{pathak2016context}, or predicting image rotation \cite{gidaris2018unsupervised}. However, pretext tasks have to be handcrafted,  and the generality of their representations is typically limited~\cite{chen2020simple}.

Hence, researchers have recently focused on contrastive learning, which emerged as a competitive and systematic method for learning effective representations without human supervision. The learned features generalize well to downstream tasks, outperform representations learned through pretext tasks, and even surpass the performance of supervised models on some tasks \cite{chen2020simple,chen2020big,chen2020improved,he2020momentum}. Multiple successful contrastive learning frameworks have been proposed, which typically differ in the way they sample negative pairs. To name a few, SimCLR \cite{chen2020simple} uses a large batch size, and samples negative pairs within each batch. The momentum-contrastive approach (MoCo) \cite{he2020momentum} leverages a moving-average encoder and a queue to generate negative samples on the fly during training.  Contrastive-Multiview-Coding \cite{tian2019contrastive} maintains a memory-bank to store features and generate negative samples. Some recent methods, like BYOL, do not rely on negative pairs \cite{chen2020exploring,grill2020bootstrap}. Instead, they use two neural networks that learn from each other to boost performance.

Past work has also reported problems with contrastive learning. It can focus on irrelevant features such as color distribution, and suppress more relevant features \cite{chen2020simple}. Past work addressed this problem by using color-distortion as a data augmentation. Also, the authors of \cite{tian2020makes} noted that when the data includes multiple types of semantics, contrastive learning may learn one type of semantics and fail to learn effective features of the other semantics (as in Figure~\ref{fig:intro}(b) where the background object information can suppress features related to digits).  They proposed a solution that learns contrastive views suitable for the desired downstream task. While they share our goal of supporting different downstream tasks, their method requires supervision since they learn their contrastive views from labeled data. In contrast, our approach is completely unsupervised. 
 
Another related work is contrastive-predictive-coding (CPC) \cite{oord2018representation,henaff2019data}. 
CPC has some similarities with \name~in that it has a predictive task that aims to reconstruct missing information. However, CPC aims to reconstruct the features of a future frame, while \name~reconstructs the raw input data. As a result,  the representation learned by CPC is not forced to contain necessary information to reconstruct the input, making it susceptible to \red{feature suppression}, just like other contrastive learning methods.

The family of auto-encoders provides  a popular framework for unsupervised representation learning using a reconstructive loss~\cite{hinton2006reducing,pu2016variational,vincent2008extracting}. It trains an encoder to generate low-dimensional latent codes that could reconstruct the entire high-dimensional inputs. There are many types of AEs, such as denoising auto-encoders \cite{vincent2008extracting}, which corrupt the input and let the latent codes reconstruct it, and variational auto-encoders \cite{pu2016variational}, which force the latent codes to follow a prior distribution. {\name} can be viewed as a special variant of the denoising auto-encoder that forces the latent codes to have a `contrastive' property regularized by a contrastive loss. As a result, the latent codes, are good not only for reconstructing the input, but also for downstream classification tasks. 


Finally, several concurrent papers published on Arxiv also used a combination contrastive and reconstructive loss \cite{dippel2021towards,jiang2020speech}. However, none of them explore the potential of this combination to solve the \red{feature suppression} problem, or provides a theoretical analysis of \red{feature suppression}. This paper is the first to demonstrate that the combination of contrastive and predictive loss can be used to avoid \red{feature suppression} and learn general representations that support multiple downstream tasks.

 \newcommand{\gL}{\mathcal{L}}
\newcommand{\NCE}{\mathcal{L}_{\mathtt{NCE}}}
\newcommand{\ENCE}{\mathcal{E}_{\mathtt{limNCE}}}
\newcommand{\E}{\mathbb{E}}
\newcommand{\pdata}{p_{\mathtt{data}}}
\newcommand{\ppos}{p_{\mathtt{pos}}}
\newcommand{\lift}{\mathcal{T}_{\sigma}}

\section{Analysis of Feature Suppression}
Before delving into formal proofs, we provide an informal description of our analysis as follows:  
\begin{Itemize}
\item [1.] At low feature dimensions, naturally the global minimum of contrastive learning loses semantic information because with  small feature dimensions, it is impossible to keep all information about the input. 
\item[2.] We prove in Corollary 2 that the global minimums of the contrastive learning loss (i.e., infoNCE) at low dimensions are local minimums at higher dimensions. Thus, even if contrastive learning uses high dimension features, it will have many local minimums, and those minimums lose semantic information about the input, i.e., they experience feature suppression.  
\item[3.] We further prove in Lemma 1 and Figure~\ref{fig:uniformity} that the value of infoNCE at such local minimums can be very close to its global minimum, making it hard to escape from such local minimums. 
\item[4.] The above three points mean that, even at high dimensions, contrastive learning is likely to get stuck in a local minimum that exhibits feature suppression. Adding a predictive loss allows the model to exit such local minimum and avoid feature suppression. This is because suppressed features lose information about the input causing the predictive loss to become large, and push the model out from such local minimum and away from feature suppression.
\end{Itemize}

\subsection{Formal Proof.}

\noindent Let  $X=\{x_i\}^n_{i=1}$ be the set of the data points. We use $\lambda_{ij}$ to indicate whether a data pair $x_i$ and $x_j$ is positive or negative. Specifically, $\lambda_{ij}=1$ indicates a positive pair while $\lambda_{ij}=0$ indicates a negative pair.
Let $Z=\{z_i\}^n_{i=1}$, where $z_i=f(x_i)=(z_i^1, \cdots, z_i^d) \in \mathbb{S}^{d-1}$, denote the learned features on the hypersphere, generated by the neural network $f$. We consider the following empirical asymptotics of the infoNCE objective function introduced in~\cite{wang2020understanding}.

\blue{
\begin{definition}[Empirical infoNCE asymptotics]
\begin{align*}
&\ENCE(Z; X, t, d) \triangleq  \\ &-\frac{1}{t n^2} \sum_{ij} \lambda_{ij} z_i^\top z_j + \frac{1}{n} \sum_{i} \log \left( \frac{1}{n} \sum_{j} e^{z_i^\top z_j / t} \right)
\end{align*}
\end{definition}
}

We are going to connect the landscape of empirical infoNCE asymptotics in the low dimension to that in the high dimension. We start by defining a {\em lifting operator} that maps a low dimensional vector to a higher dimension. 
\begin{definition}[Lifting operator] 
A lifting operator $\lift$ parameterized by an indexing function $\sigma$ maps a $d_1$-dimensional vector to dimension $d_2$ ($d_2 > d_1$). Its parameter $\sigma$ is a permutation of length $d_2$. Given a $d_1$-dimensional vector $z$, the lifting operator maps it to a $d_2$-dimensional vector $\tilde{z} = \lift(z)$ by the following rules: $\tilde{z}^t = z^{\sigma(t)}$ if $\sigma(t) \leq d_1$, otherwise $\tilde{z}^t = 0$.
\end{definition}

\noindent With a slight abuse of the notation, we allow the lifting operator to map a {\em set} of low dimensional vectors to higher dimension, i.e. $\lift(\{z_i\}) = \{ \lift(z_i) \}$. We further allow the lifting operator to map a function $f$ of lower dimension to higher dimension, i.e., $\lift(f)(x) = \lift(f(x))$. Note that $\lift$ is a linear operator. We highlight several useful properties of $\lift$:

\blue{
\begin{lemma}[Value Invariance]\label{lem:value}
The value of the empirical infoNCE asymptotics is invariant under the lifting operation. Formally, consider any lifting operator $\lift$ from the dimension $d_1$ to the dimension $d_2$. We have 
\begin{align*}
\ENCE(\lift(Z); X, t, d_2) = \ENCE(Z; X, t, d_1) 
\end{align*}
\end{lemma}
}

\begin{proof}
Following the definition of $\lift$, $\forall z_i, z_j$, $z_i^\top z_j = \lift(z_i)^\top \lift(z_j)$. Therefore, $\ENCE(\lift(Z); X, t, d_2) = \ENCE(Z; X, t, d_1)$.
\end{proof}

\blue{
\begin{lemma}[Gradient Equivariance]\label{lem:gradient}
The gradient of the empirical infoNCE asymptotics is equivariant under the lifting operation. Formally, consider any lifting operator $\lift$ from the dimension $d_1$ to the dimension $d_2$. We have 
\resizebox{\linewidth}{!}{
  \begin{minipage}{\linewidth}
\begin{align*}
\nabla_{\tilde{z_k}}\ENCE(\lift(Z); X, t, d_2) = \lift \left( \nabla_{z_k} \ENCE(Z; X, t, d_1) \right)
\end{align*}
\end{minipage}
}
\end{lemma}
\begin{proof}
$$$$
\resizebox{\linewidth}{!}{
  \begin{minipage}{\linewidth}
\vspace{-15pt}
\begin{align*}
&\nabla_{z_k}\ENCE(Z; X, t, d_1)
\\ & \triangleq  \nabla_{z_k}\Big(-\frac{1}{t n^2} \sum_{ij} \lambda_{ij} z_i^\top z_j + \frac{1}{n} \sum_{i} \log \left( \frac{1}{n} \sum_{j} e^{z_i^\top z_j / t} \right)\Big)
\\ & = -\frac{1}{t n^2} (\sum_{i\neq k} \lambda_{ki} z_i + \sum_{i\neq k} \lambda_{ik} z_i + 2\lambda_{kk}z_k)
\\ & + \frac{1}{t n} \frac{2z_k e^{z_k^\top z_k / t} + \sum_{j\neq k} z_j e^{z_k^\top z_j / t}}{\sum_{j} e^{z_k^\top z_j / t}}
+ \frac{1}{t n} \sum_{i \neq k} \frac{z_i e^{z_i^\top z_k / t}}{\sum_{j} e^{z_i^\top z_{j} / t}}
\\ & = -\frac{1}{t n^2} (\sum_{i} \lambda_{ki} z_i + \sum_{i} \lambda_{ik} z_i)
\\ & + \frac{1}{t n} \frac{z_k e^{z_k^\top z_k / t} + \sum_{j} z_j e^{z_k^\top z_j / t}}{\sum_{j} e^{z_k^\top z_j / t}}
+ \frac{1}{t n} \sum_{i \neq k} \frac{z_i e^{z_i^\top z_k / t}}{\sum_{j} e^{z_i^\top z_{j} / t}}.
\end{align*}
\end{minipage}
}
Since $\lift$ is a linear operator,
\begin{align*}
&\lift \left( \nabla_{z_k} \ENCE(Z; X, t, d_1) \right)
\\ & = -\frac{1}{t n^2} (\sum_{i} \lambda_{ki} \tilde{z_i} + \sum_{i} \lambda_{ik} \tilde{z_i})
\\ & + \frac{1}{t n} \frac{\tilde{z_k} e^{z_k^\top z_k / t} + \sum_{j} \tilde{z_j} e^{z_k^\top z_j / t}}{\sum_{j} e^{z_k^\top z_j / t}}
+ \frac{1}{t n} \sum_{i \neq k} \frac{\tilde{z_i} e^{z_i^\top z_k / t}}{\sum_{j} e^{z_i^\top z_{j} / t}}
\\ & = -\frac{1}{t n^2} (\sum_{i} \lambda_{ki} \tilde{z_i} + \sum_{i} \lambda_{ik} \tilde{z_i})
\\ & + \frac{1}{t n} \frac{ \tilde{z_k} e^{\tilde{z_k}^\top \tilde{z_k} / t} + \sum_{j} \tilde{z_j} e^{\tilde{z_k}^\top \tilde{z_j} / t}}{\sum_{j} e^{\tilde{z_k}^\top \tilde{z_j} / t}}
+ \frac{1}{t n} \sum_{i \neq k} \frac{\tilde{z_i} e^{\tilde{z_i}^\top \tilde{z_k} / t}}{\sum_{j} e^{\tilde{z_i}^\top \tilde{z_{j}} / t}}
\\ & = \nabla_{\tilde{z_k}}\ENCE(\lift(Z); X, t, d_2)  
\end{align*}
where $\tilde{z_k} = \lift(z_k)$. The second equality comes from the fact that $\forall z_i, z_j$, $z_i^\top z_j = \lift(z_i)^\top \lift(z_j)$. Thus $\nabla_{\tilde{z_k}}\ENCE(\lift(Z); X, t, d_2)$ equals $\lift \left( \nabla_{z_k} \ENCE(Z; X, t, d_1) \right)$, for all $z_{k}\in Z$.
\end{proof}
}

\begin{corollary}\label{cor:stationary}
	For any lifting operator $\lift$, if $\hat{Z} = \{\hat{z_i}\}$ is a stationary point of $\ENCE(Z; X, t, d_1)$, then $\lift(\hat{Z})$ is a stationary point of $\ENCE(Z; X, t, d_2)$.
\end{corollary}
\begin{proof}
The proof is in the supplemental material. 
\end{proof}

\begin{corollary}\label{cor:saddle}
	For any lifting operator $\lift$, if $\hat{Z} = \{\hat{z_i}\}$ is a global minimum of $\ENCE(Z; X, t, d_1)$ with a positive definite Hessian matrix, then $\lift(\hat{Z})$ is a saddle point or a local minimum of $\ENCE(Z; X, t, d_2)$.
\end{corollary}

\begin{proof}
The proof is in the supplemental material. 
\end{proof}

\begin{figure}
\begin{center}
\includegraphics[width=0.9\linewidth]{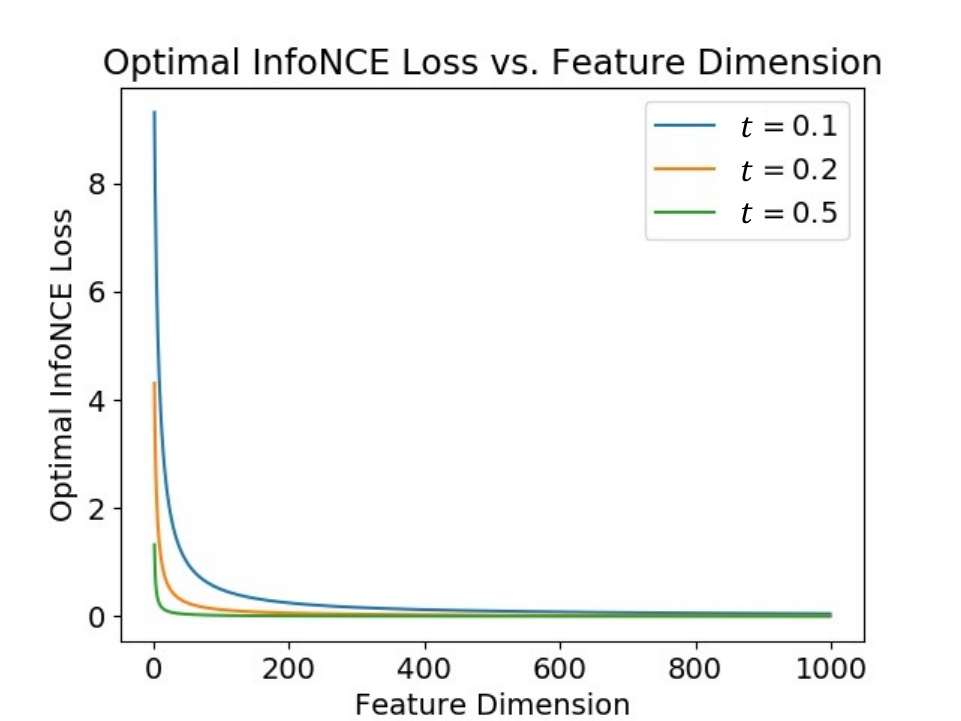}
\vspace{-10pt}
\end{center}
\caption{\footnotesize{Optimal infoNCE loss vs. different output feature dimension $d$ and temperature $t$.}}\label{fig:uniformity}
\vspace{-5pt}
\end{figure}

With Corollary \ref{cor:saddle}, we can explain why contrastive learning can suffer from \red{feature suppression}. Suppose $f$ is a network that achieves the global minimum of $\ENCE(Z; X, t, d_1)$. When $d_1$ is relatively small (e.g., <100 for images), $f$ must lose some information about the input, i.e., \red{suppress feature}.  From Corollary \ref{cor:saddle}, $\lift(f)$ is a saddle point or a local minimum of $\ENCE(Z; X, t, d_2)$ where $d_2>d_1$ and $\lift(f)$ carries no more information than $f$. Therefore, for any dimension $d>1$, there exists saddle point/local minimum of $\ENCE(Z; X, t, d)$ which \red{suppresses features}.


Furthermore, the value of the aforementioned saddle point/local minimum of $\ENCE(Z; X, t, d)$ is quite close to that of the global minimum. This is because the optimal value of $\ENCE(Z; X, t, d)$ converges quickly as $d$ increases. Figure \ref{fig:uniformity} shows the curve of $\log _0F_1(;d;\frac{1}{4t^2})$, which is the optimal value of the infoNCE loss \cite{wang2020unsupervised}. As shown in the figure, the curve essentially converges when $d>200$. Therefore, $\lift(f)$ can be a saddle point/local minimum of $\ENCE(Z; X, t, d_2)$, and its value can also be quite close to that of the global minimum, making it hard to escape from such local minimum. So effectively one can achieve a value pretty close to the global minimum by \red{suppressing features}, and stay at that saddle point being unable to escape. 
This motivates our solution, which adds a predictive loss to force the model out from such local minimums that suppress features. 
 \section{\red{Predictive} Contrastive Learning (\name)}
\red{Predictive} contrastive learning (\name) is a framework for self-supervised representation learning. It aims to learn representations that are robust to \red{feature suppression}, and capable of supporting multiple diverse downstream tasks. 

\begin{figure*}
\begin{center}
\includegraphics[width=0.8\linewidth]{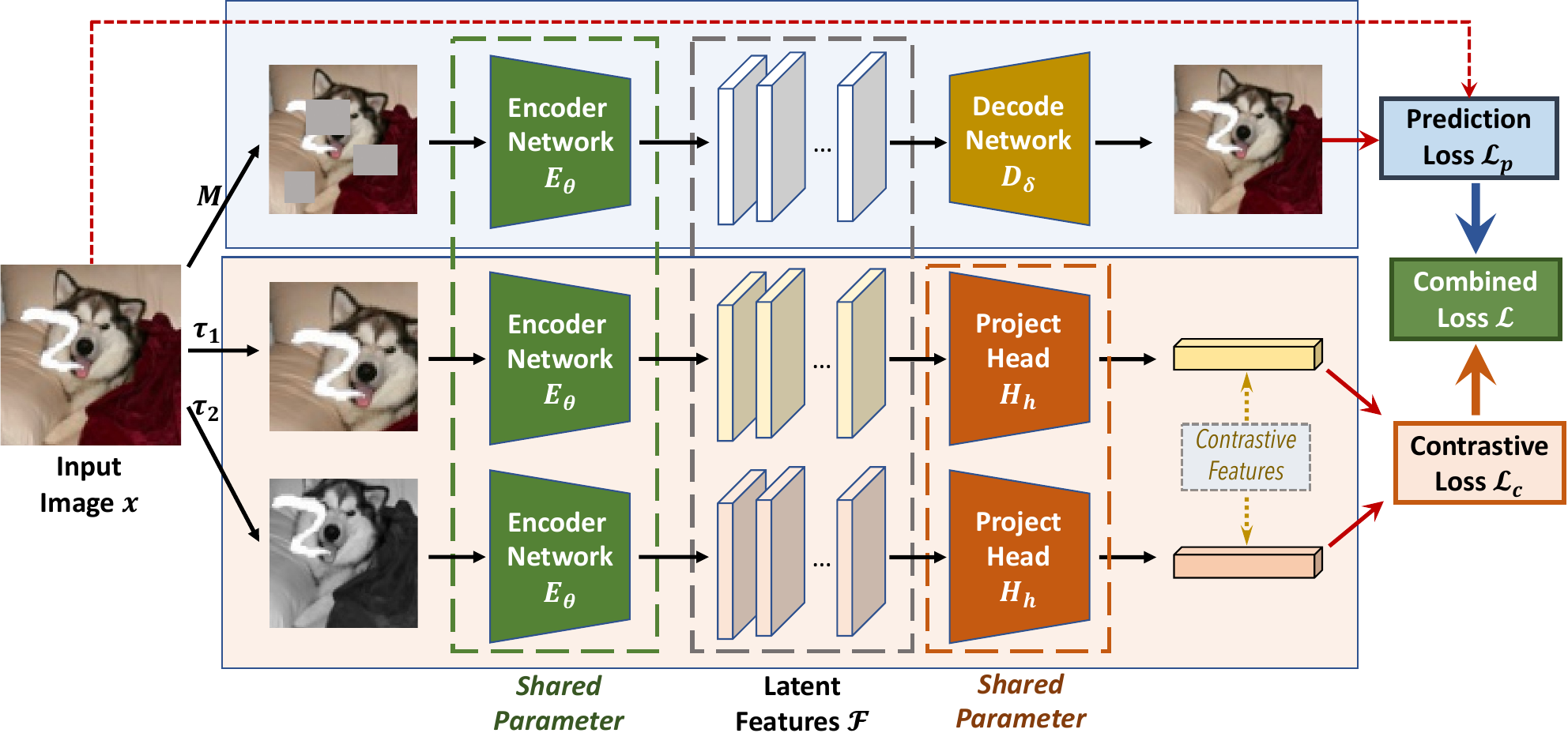}
\end{center}
\caption{\footnotesize{Illustration of the \name~framework. \name\ has two branches: 1) a \red{predictive} branch, illustrated in the blue box, which ensures that the representation has enough information to restore missing patches in the input, and 2) a contrastive branch, illustrated in the orange box, which ensures that the representation keeps positive samples close to each other and away from negative samples.}}\label{fig:framework}
\vspace{-5pt}
\end{figure*}

The idea underlying \name\ is as follows: \red{feature suppression} is harmful because the representation loses important information that was available in the input. Thus, to counter \red{feature suppression}, \name\ uses a prediction loss to ensure that the representation can restore the input, i.e., the features have the information available at the input. Yet, keeping all information in the features is not enough; the input already has all information. 
By adding a contrastive loss, \name\ reorganizes the information in the feature space to make it amenable to downstream classification, i.e., samples that have similar attributes/objects are closer to each other than samples that have different attributes/objects. Figure \ref{fig:framework} shows the \name\ framework which has two branches: a contrastive branch and a predictive branch.



{\bf (a) Contrastive Branch:} The contrastive branch is illustrated in the orange box in Figure~\ref{fig:framework}. Here, we use SimCLR as an example to demonstrate the basic idea. However, this contrastive branch can be easily adapted to any contrastive learning method such as CPC, MoCo, and BYOL. For each image, we first generate a pair of positive samples by using two random augmentations $\tau_1$ and $\tau_2$, then we forward the two augmented inputs separately to the encoder $E$, parameterized by $\theta$ and a multi-layer nonlinear projection head $H$ parameterized by $h$ to get the latent representations $z_{1}$ and $z_{2}$ for these two positive samples. We use the commonly used InfoNCE loss \cite{chen2020simple} as the contrastive loss $\mathcal{L}_{c}$. Namely, for a batch of $N$ different input images $x_{i}, i=1,...,N $,
$$
\mathcal{L}_{c} = -\sum_{i=1}^N \log \sum \frac{\exp\big(\text{sim}(z_{2i}, z_{2i+1})/t\big)}{\sum_{k=1}^{2N} \mathbbm{1}_{k\neq 2i}\exp\big(\text{sim}(z_{2i}, z_k)/t\big)},
$$
where $\text{sim}(u,v)=u^{T}v / (\|u\|_2 \|v\|_2)$ denotes the dot product between the normalized $u$ and $v$ (i.e., cosine similarity), $t \in \mathbb{R}^{+}$ is a scalar temperature parameter, and $z_{2i},z_{2i+1}$ are the encoded features of positive pairs generated from $x_{i}$, i.e., $z_{2i} = H_h(E_{\theta}(\tau_1(x_{i})))$ and $z_{2i+1} = H_h(E_{\theta}(\tau_2(x_{i})))$.

{\textbf{(b) \red{Predictive} Branch:}} 
To choose a proper \red{predictive} task, we need to consider two aspects: its ability to summarize and abstract the input, and its applicability to different datasets and tasks. In fact, many self-supervised learning tasks, such as Auto-encoder, Colorization and Inpainting, are \red{predictive} since they all aim to restore the input. But, those tasks do not have the same ability to both retain and abstract information.  For example, inpainting is a stronger \red{predictive} task than autoencoding in terms of its ability to both abstract and retain information. Thus, although both of them would help in strengthening contrastive learning against \red{feature suppression}, inpainting is likely to provide more gains. 

Another issue to consider is the applicability of the chosen task to various datasets. For example, colorization is applicable only to colorful RGB datasets, but not to grey-scale datasets such as MNIST or medical image datasets. In contrast, a task like inpainting is easier to translate across different datasets. 

Given the above considerations, we adopt inpainting as the default \red{predictive} task.  In the supplemental material, we compare various tasks and show that while they all improve performance, inpainting delivers higher gains.

Figure \ref{fig:framework} shows how \name~uses the inpainting task, where given an input image $x$, we first randomly mask several patches to get the masked input $M(x)$. Then the masked input is passed through an encoder network $E$ with parameter $\theta$, and a decoder network $D$, with parameter $\delta$, to obtain the reconstruction result $ D_{\delta}(E_{\theta}(M(x)))$. The \red{prediction} loss $\mathcal{L}_{p}$ is defined as the reconstruction error between the original input $x$ and the reconstructed one $D_{\delta}(E_{\theta}(M(x)))$:
$$
\mathcal{L}_{p} = ||D_{\delta}(E_{\theta}(M(x)))-x||_2.
$$

{\bf (c) Training Procedure:}
We have empirically found that it is better to train the model in two phases. In the first phase, only the \red{predictive} branch is trained. In the second phase, both branches are trained together. In this latter case,
the overall training loss is the combination of the \red{prediction} loss and the contrastive loss, i.e., $\mathcal{L} = \mathcal{L}_{c} + \lambda  \cdot \mathcal{L}_{p}$.  We set $\lambda=10$ for all experiments. We also include results with different $\lambda$ in the supplemental material.

\blue{
{\bf (d) \name~Avoids \red{Feature Suppression}:}
With a combination of the prediction loss and the contrastive loss, \name~is capable of escaping the aforementioned local minimum/saddle points of infoNCE loss where only partial semantics are learned. This is because learning only part of the semantics can result in very high prediction loss. For example, if the network learns only semantics related to the background object but ignores the digit (Figure \ref{fig:framework}), all pixels related to the digit are likely to be predicted incorrectly, introducing large gradients that force the model out of the saddle point. 
}
%
%
%
%

 \section{Experiments}
\label{sec:exp}

\begin{table*}[t]
\centering
\caption{\footnotesize{Performance on ImageNet with progressive removal of data augmentations for different self-supervised learning techniques. The baseline corresponds to the original set of augmentations used in SimCLR and MoCo: random flip, random resized crop, color distortion, and random Gaussian blur.}}
\vspace{-10pt}
\label{tab:imgnet}
\begin{threeparttable}
\footnotesize{(a) ImageNet \textsc{Top-1} accuracy and its \textsc{Drop} w.r.t. inclusion of all augmentations.}\vspace{0pt}
\resizebox{0.8\textwidth}{!}{
\begin{tabular}{l|cc|cc|cc|cc|c} 
\toprule[1.5pt]
Method    & \multicolumn{2}{c|}{\red{Inpainting}}          & \multicolumn{2}{c|}{SimCLR} & \multicolumn{2}{c|}{MoCo}  & \multicolumn{2}{c|}{\bf \name (ours)}     & \multirow{2}{*}{\bf\textsc{Improve}}        \\ 
\cmidrule{1-9}
\bf\textsc{Metric}              & \bf\textsc{Top-1} & \bf\textsc{Drop}                & \bf\textsc{Top-1} & \bf\textsc{Drop}              &\bf \textsc{Top-1} &\bf\textsc{Drop}              & \bf\textsc{Top-1}         & \bf\textsc{Drop}          &                         \\ 
\midrule\midrule
Baseline       &  43.7  & /       & 67.9  & /                   & \bf  71.1	  & /                                &71.0          & /             & \textcolor{lightblue}{\textbf{-0.1}}                       \\ 
\midrule
Remove flip    & 43.4  & -0.3       & 67.3  & -0.6                & 70.6  & -0.5                          & \bf  70.8          & -0.2          & \textcolor{darkgreen}{\textbf{+0.2}}  \\
Remove blur    & 43.6  & -0.1      & 65.2  & -2.7                & 69.7  & -1.4                           &  \bf 70.6 & -0.4 & \textcolor{darkgreen}{\textbf{+0.9}} \\
Crop  color only & 43.2  & -0.5     & 64.2  & -3.7                & 69.5  & -1.6                          &  \bf  70.1           & -0.9           & \textcolor{darkgreen}{\textbf{+0.6}} \\
Remove color distort   & 43.5  & -0.2      & 45.7  &   -22.2     & 60.4  & -10.7                         & \bf  65.9 & -5.1 & \textcolor{darkgreen}{\textbf{+5.5}}   \\
Crop  blur only  & 42.8  & -0.9    & 41.7  & -26.2               & 59.8  & -11.3                         & \bf  65.1           & -5.9           & \textcolor{darkgreen}{\textbf{+5.3}}  \\
Crop flip only  & 43.3  & -0.4     & 40.2  & -27.7               & 59.4  & -11.7                        &  \bf 64.6 & -6.4 & \textcolor{darkgreen}{\textbf{+5.2}}  \\
Crop only     & 42.7  & -1.0      & 40.3  & -27.6                & 59.0  & -12.1                          &  \bf 64.1 & -6.9 & \textcolor{darkgreen}{\textbf{+5.1}} \\
\bottomrule[1.5pt]
\end{tabular}
}
\end{threeparttable}
\begin{threeparttable}
\footnotesize{(b) ImageNet \textsc{Top-5} accuracy and its \textsc{Drop} w.r.t.  inclusion of all augmentations.}
\resizebox{0.8\textwidth}{!}{
\begin{tabular}{l|cc|cc|cc|cc|c} 
\toprule[1.5pt]
 Method     & \multicolumn{2}{c|}{\red{Inpainting}}         & \multicolumn{2}{c|}{SimCLR} & \multicolumn{2}{c|}{MoCo}  & \multicolumn{2}{c|}{\bf \name (ours)}     & \multirow{2}{*}{\bf\textsc{Improve}}        \\ 
\cmidrule{1-9}
\bf\textsc{Metric}              & \bf\textsc{Top-5} & \bf\textsc{Drop}                & \bf\textsc{Top-5} & \bf\textsc{Drop}              &\bf \textsc{Top-5} &\bf\textsc{Drop}              & \bf\textsc{Top-5}         & \bf\textsc{Drop}                                 \\ 
\midrule\midrule
Baseline     & 68.3  & /        & 88.5  & /                   &  \bf 90.1  & /                                 & 90.0          & /             & \textcolor{lightblue}{\textbf{-0.1}}                       \\ 
\midrule
Remove flip    & 67.9  & -0.4      & 88.2  & -0.3                & 89.9  & -0.2                           &   \bf  89.9        & -0.1          & \textcolor{darkgreen}{\textbf{+0.0}}         \\
Remove blur   & 68.1  & -0.2        & 86.6  & -1.9                & 89.7  & -0.4                          & \bf  89.8 & -0.2 & \textcolor{darkgreen}{\textbf{+0.1}}  \\
Crop  color only  & 67.8  & -0.5    & 86.2  & -2.3                & 89.6  & -0.5                         &  \bf 89.7           &-0.3           & \textcolor{darkgreen}{\textbf{+0.1}} \\
Remove color distort    & 68.0  &   -0.3    & 70.6  &    -17.9         & 84.2  & -5.9                &  \bf 88.3 & -1.7 & \textcolor{darkgreen}{\textbf{+4.1}}  \\
Crop  blur only   & 67.4  & -0.9    & 66.4  &      -22.1     & 83.1  & -7.0                  &  \bf 88.0          & -2.0          & \textcolor{darkgreen}{\textbf{+4.9}}   \\
Crop flip only   & 67.7  & -0.6    & 64.8  &   -23.7         & 82.0  & -8.1                         &  \bf 87.7 & -2.3 & \textcolor{darkgreen}{\textbf{+5.7}} \\
Crop only  & 67.4  & -0.9          & 64.8  & -23.7               & 81.6  & -8.5                &  \bf 87.6 & -2.4 & \textcolor{darkgreen}{\textbf{+6.0}} \\
\bottomrule[1.5pt]
\end{tabular}
}
\end{threeparttable}


\caption{\footnotesize{Performance on MPII for the downstream task of human pose estimation. $\uparrow$ indicates the larger the value, the better the performance.}}
\label{tab:result-pose}
\begin{threeparttable}
\resizebox{0.8\textwidth}{!}{
\begin{tabular}{c|c|c|c|c|c|c|c|c|c} 
\toprule[1.5pt]
\multicolumn{2}{c|}{\textsc{Metric}} & Head$^\uparrow$ & Shoulder$^\uparrow$ & Elbow$^\uparrow$ & Wrist$^\uparrow$ & Hip$^\uparrow$ & Knee$^\uparrow$ & Ankle$^\uparrow$ & PCKh$^\uparrow$ \\ 
\midrule
\midrule
\multirow{5}{*}{\begin{tabular}[c]{@{}c@{}}\textsc{Fixed}\\\textsc{feature}\\\textsc{extractor} \end{tabular}} 

&SimCLR & 78.4 & 74.6 & 56.7 & 45.2 & 61.8 & 51.3 & 47.1 & 60.8 \\
&MoCo & 79.2 & 75.1 & 57.4 & 45.9 & 62.4 & 52.0 & 47.6 & 61.4 \\

&CPC & 78.0 & 74.3 & 56.0 & 44.8 & 61.2 & 51.4 & 46.5 & 60.3 \\
&BYOL & 79.1 & 75.0 & 57.1 & 46.0 & 62.4 & 52.2 & 47.7 & 61.4 \\
\cmidrule{2-10}
&\bf \name~(ours)                                                                                      & \bf 85.7 & \bf 78.8 & \bf 61.7 & \bf 51.3 & \bf 64.4 & \bf 55.6 & \bf 49.2 & \bf 65.1                                                                                            \\ 
&\bf \textsc{Improvements}                                                                                      
& \bf \textcolor{darkgreen}{\textbf{+6.5}}
& \bf \textcolor{darkgreen}{\textbf{+3.7}} 
& \bf \textcolor{darkgreen}{\textbf{+4.3}} 
& \bf \textcolor{darkgreen}{\textbf{+5.3}} 
& \bf \textcolor{darkgreen}{\textbf{+2.0}} 
& \bf \textcolor{darkgreen}{\textbf{+3.4}} 
& \bf \textcolor{darkgreen}{\textbf{+1.5}}
& \bf \textcolor{darkgreen}{\textbf{+3.7}}                                                                                          \\ 
\midrule
\midrule
\multirow{5}{*}{\begin{tabular}[c]{@{}c@{}}\textsc{Fine-}\\\textsc{tuning} \end{tabular}} 
& SimCLR & 96.2 & 94.7 & 87.3 & 81.2 & 87.5 & 81.0 & 77.2 & 87.1 \\
& MoCo & 95.9 & 94.7 & 87.5 & 81.6 & 87.4 & 81.7 & 76.9 & 87.2 \\
& CPC & 96.0 & 94.5 & 87.0 & 81.1 & 87.3 & 80.8 & 77.0 & 87.0 \\
& BYOL & 96.2 & 94.8 & 87.5 & 81.4 & 87.6 & 81.5 & 77.0 & 87.2 \\
\cmidrule{2-10}
&\bf \name~(ours)                                                                                     & \bf 96.3 & \bf 94.9 & \bf 88.1 & \bf 82.3 & \bf 87.9 & \bf 82.8 & \bf 77.8 & \bf 87.8                                                                                                                                                                        \\
&\bf \textsc{Improvements}                                                                                      
& \bf \textcolor{darkgreen}{\textbf{+0.1}} 
& \bf \textcolor{darkgreen}{\textbf{+0.1}}  
& \bf \textcolor{darkgreen}{\textbf{+0.6}} 
& \bf \textcolor{darkgreen}{\textbf{+0.7}}
& \bf \textcolor{darkgreen}{\textbf{+0.3}} 
& \bf \textcolor{darkgreen}{\textbf{+1.1}} 
& \bf \textcolor{darkgreen}{\textbf{+0.6}} 
& \bf \textcolor{darkgreen}{\textbf{+0.6}}                                                                                            \\ 

\midrule
\midrule
\multicolumn{2}{c|}{\textsc{Supervised}}     & 96.3 & 95.1 & 87.9 & 82.2 & 87.8 & 82.7 & 77.8 & 87.7\\

\bottomrule[1.5pt]
\end{tabular}}
\end{threeparttable}
\vspace{-15pt}
\end{table*}

\noindent\textbf{Baselines.}
We use state-of-the-art contrastive learning methods as baselines, including SimCLR \cite{chen2020simple}, MoCo \cite{chen2020improved}, CPC \cite{henaff2019data} and BYOL \cite{grill2020bootstrap}. The same network structure, batch size, and training epochs are used for all baselines and \name's contrastive branch. For the contrastive branch of \name, we apply the same training scheme as MoCo. \red{\name~uses the \red{predictive} branch only for training. During inference it uses only the encoder, which is shared with the contrastive branch. Thus, the evaluation of \name~uses exactly the same number of parameters as the baselines.}

\noindent\textbf{Datasets.} We experiment with the following datasets:
\begin{Itemize}
\item 
\textbf{ImageNet:} ImageNet\cite{deng2009imagenet} (CC  BY  2.0) is a widely used image classification benchmark which contains 1.28M images in 1000 different categories. It is a standard benchmark to evaluate self-supervised learning methods \cite{chen2020improved,chen2020simple,grill2020bootstrap}.
\item
\textbf{MPII:} MPII~\cite{andriluka20142d} (the Simplified BSD License)  is one of the most common datasets for the task of human pose estimation. It contains images of everyday human activities.
\item
\textbf{FairFace:} FairFace \cite{karkkainen2019fairface} (CC BY 4.0.) is a face attribute classification dataset, where each image contains multiple semantics including gender, age, and ethnicity. 
\item
\textbf{Colorful-Moving-MNIST:} This is a synthetic dataset used by~\cite{tian2020makes} to highlight the \red{feature suppression} problem. It is constructed by assigning each digit from MNIST a background object image selected randomly from STL-10. It supports two downstream tasks: digit and background classification.
\end{Itemize}


\begin{table*}
\begin{tabular}{cc}

\begin{minipage}{.52\linewidth}
\centering
\caption{\footnotesize{Performance on FairFace with different unsupervised learning methods. The models are evaluated on downstream tasks of age, gender and ethnicity classification.}}
\label{tab:result-face}
\begin{threeparttable}
\resizebox{.95\textwidth}{!}{
\begin{tabular}{c|c|c|c|c} 
\toprule[1.5pt]
\multicolumn{2}{c|}{\textsc{Metric}}                                                                                  & \begin{tabular}[c]{@{}c@{}}\textsc{Age cls} \\ \textsc{Acc.} (\%) \end{tabular} & \begin{tabular}[c]{@{}c@{}}\textsc{Gender cls} \\ \textsc{Acc.} (\%) \end{tabular} & \begin{tabular}[c]{@{}c@{}}\textsc{Ethn. cls} \\ \textsc{Acc.} (\%) \end{tabular} \\ 
\midrule\midrule
\multirow{6}{*}{\begin{tabular}[c]{@{}c@{}}\textsc{Fixed}\\\textsc{feature}\\\textsc{extractor} \end{tabular}} &SimCLR                   & 43.9  & 78.1  & 61.7                                                                  \\
&MoCo                      & 44.5  & 78.6  & 61.9                                                                  \\
&CPC                                                                                                         & 43.5  & 76.2  & 61.0                                                                  \\ 
&BYOL & 44.3  & 78.6  & \textbf{62.3}  \\
\cmidrule{2-5}
&\bf{\name~(ours)}  & \bf{50.0}  & \bf{87.2}  & 61.2     \\ 
&\bf \textsc{Improvement}                                                                                      & \bf \textcolor{darkgreen}{\textbf{+5.7}}  & \textcolor{darkgreen}{\textbf{+8.6}}  & \textcolor{lightblue}{\textbf{-1.1}} \\ 
\midrule\midrule
\multirow{6}{*}{\begin{tabular}[c]{@{}c@{}}\textsc{Fine-}\\\textsc{tuning} \end{tabular}}
& SimCLR  &54.3 & 91.1 & 69.1   \\
& MoCo & 54.7 & 91.3 & 69.2  \\
&CPC  & 54.2 & 91.0 & 68.8    \\ 
&BYOL & 54.6 &91.5 & \textbf{69.3} \\
\cmidrule{2-5}
&\bf{\name~(ours)}                                                                                    &  \textbf{55.3}                                                                  &  \textbf{92.3}                 & 69.0                                                 \\ 
&\bf \textsc{Improvement}                                                                                     & \bf \textcolor{darkgreen}{\textbf{+0.6}}      & \bf \textcolor{darkgreen}{\textbf{+0.8}}                                                         & \textcolor{lightblue}{\textbf{-0.3}}                                                             \\ 
\midrule\midrule
\multicolumn{2}{c|}{\textsc{Supervised} on \textsc{Age}}                                                                      & 55.5 & 78.8  & 45.1                                                                 \\
\multicolumn{2}{c|}{\textsc{Supervised} on \textsc{gender}}                                                                        & 43.3 & 92.5  & 45.4                                                                 \\
\multicolumn{2}{c|}{\textsc{Supervised} on \textsc{Ethn.}}                                                                & 42.1 & 76.8  & 69.4                                                                \\
\multicolumn{2}{c|}{\textsc{Supervised} on \textsc{All}}                                                                & 54.8 & 91.9  &  68.8                                                                \\
\bottomrule[1.5pt]
\end{tabular}
}
\end{threeparttable}
\end{minipage}
&
\begin{minipage}{.46\linewidth}
\centering
\caption{\footnotesize{Performance on Colorful-Moving-MNIST under different unsupervised methods. The models are evaluated on the downstream tasks of digit classification and background object classification.}}
\label{tab:result-toy}
\begin{threeparttable}
\resizebox{0.9\textwidth}{!}{
\begin{tabular}{c|c|c|c} 
\toprule[1.5pt]
\multicolumn{2}{c|}{\textsc{Metric}}                                                                                  & \begin{tabular}[c]{@{}c@{}}\textsc{Digit cls} \\ \textsc{Acc.} (\%) \end{tabular} & \begin{tabular}[c]{@{}c@{}}\textsc{Bkgd cls} \\ \textsc{Acc.} (\%) \end{tabular}  \\ 
\midrule\midrule
\multirow{6}{*}{\begin{tabular}[c]{@{}c@{}}\textsc{Fixed}\\\textsc{feature}\\\textsc{extractor} \end{tabular}} &SimCLR                    & 14.9                                                                  & 47.3                                                                  \\
&MoCo                                                                                                         & 15.7                                                                  & 48.5                                                                  \\
&CPC                                                                                                          & 15.8                                                                  & 35.2                                                                  \\ 
&BYOL & 15.5 & \textbf{49.0} \\
\cmidrule{2-4}
&\bf{\name~(ours)}                                                                                      & \bf 88.3                                                                  & 46.5                                                                  \\ 
&\bf \textsc{Improvement}                                                                                      & \bf \textcolor{darkgreen}{\textbf{+72.5}}                                                                & \textcolor{lightblue}{\textbf{-2.5}} \\ 
\midrule\midrule
\multirow{6}{*}{\begin{tabular}[c]{@{}c@{}}\textsc{Fine-}\\\textsc{tuning} \end{tabular}} & SimCLR                               & 92.4                                                                   & 54.8                                                                  \\
& MoCo                                                                                                         & 92.7                                                                   & 54.9                                                                  \\
&CPC                                                                                                      & 92.3                                                                   & 54.7                                                                  \\ 
&BYOL & 92.7 & \textbf{54.9} \\
\cmidrule{2-4}
&\bf{\name~(ours)}                                                                                    & \bf 93.3                                                                   & 54.7                                                                  \\ 
&\bf \textsc{Improvement}                                                                                     & \bf \textcolor{darkgreen}{\textbf{+0.6}}                                                               & \textcolor{lightblue}{\textbf{-0.2}}                                                             \\ 
\midrule\midrule
\multicolumn{2}{c|}{\textsc{Supervised} on \textsc{digit}}                                                                      & 96.1                                                                   & 11.4                                                                  \\
\multicolumn{2}{c|}{\textsc{Supervised} on \textsc{bkgd}}                                                                       & 12.9                                                                  & 56.7                                                                  \\
\multicolumn{2}{c|}{\textsc{Supervised} on \textsc{digit} \&  \textsc{bkgd}}                                                                & 93.0                                                                   & 54.5                                                                  \\
\bottomrule[1.5pt]
\end{tabular}
}
\end{threeparttable}


\end{minipage}
\end{tabular}

\end{table*}
\subsection{Results}
\label{sec:rgb}

\begin{figure*}[h]
\begin{center}
\vspace{5pt}
\begin{tabular}{ccccc}
\includegraphics[width=0.23\linewidth]{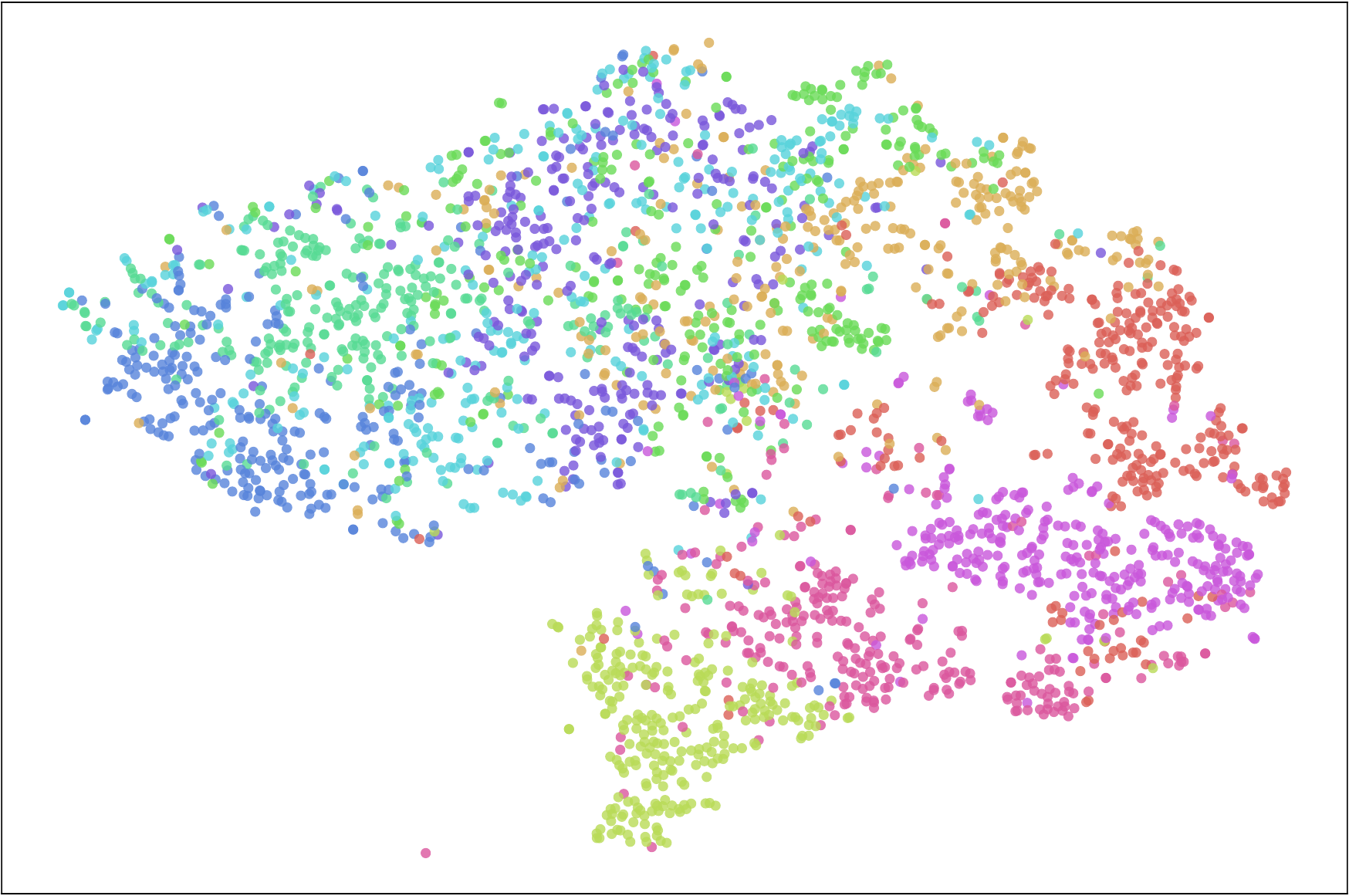} & \hspace*{-0.18in}
\includegraphics[width=0.23\linewidth]{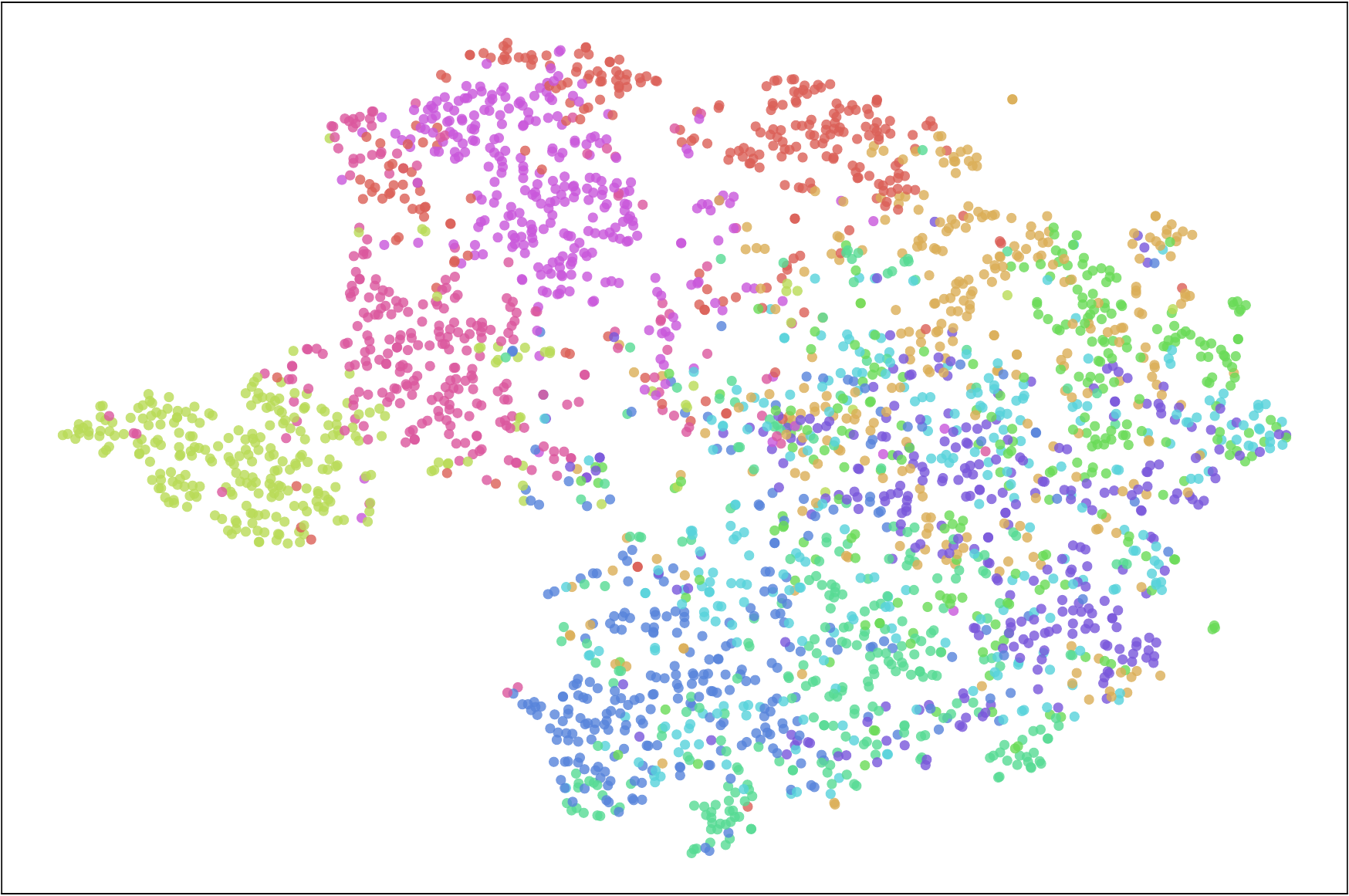}  & \hspace*{-0.06in}
&
\hspace*{-0.06in}
\includegraphics[width=0.23\linewidth]{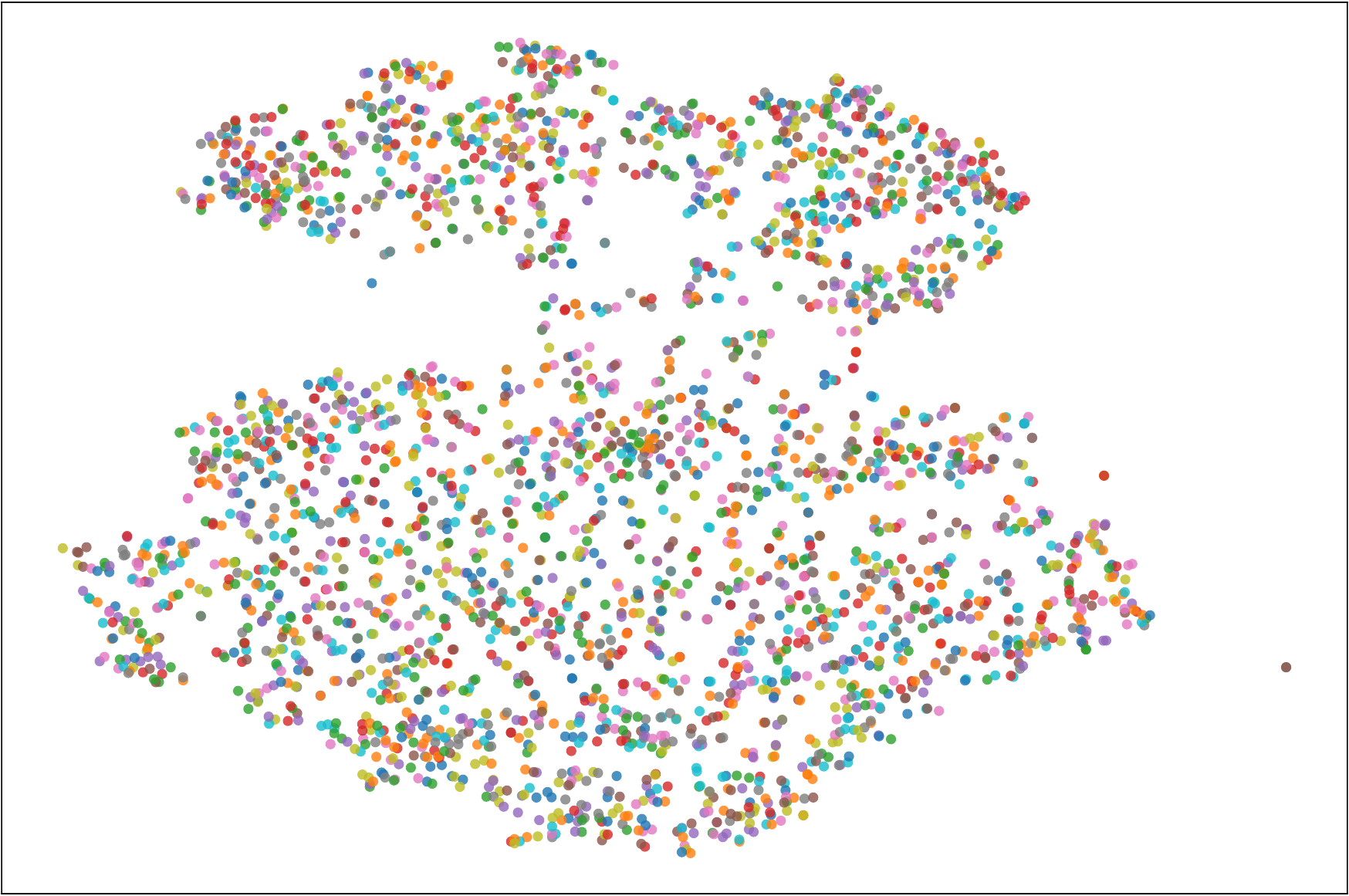}  & \hspace*{-0.18in}
\includegraphics[width=0.23\linewidth]{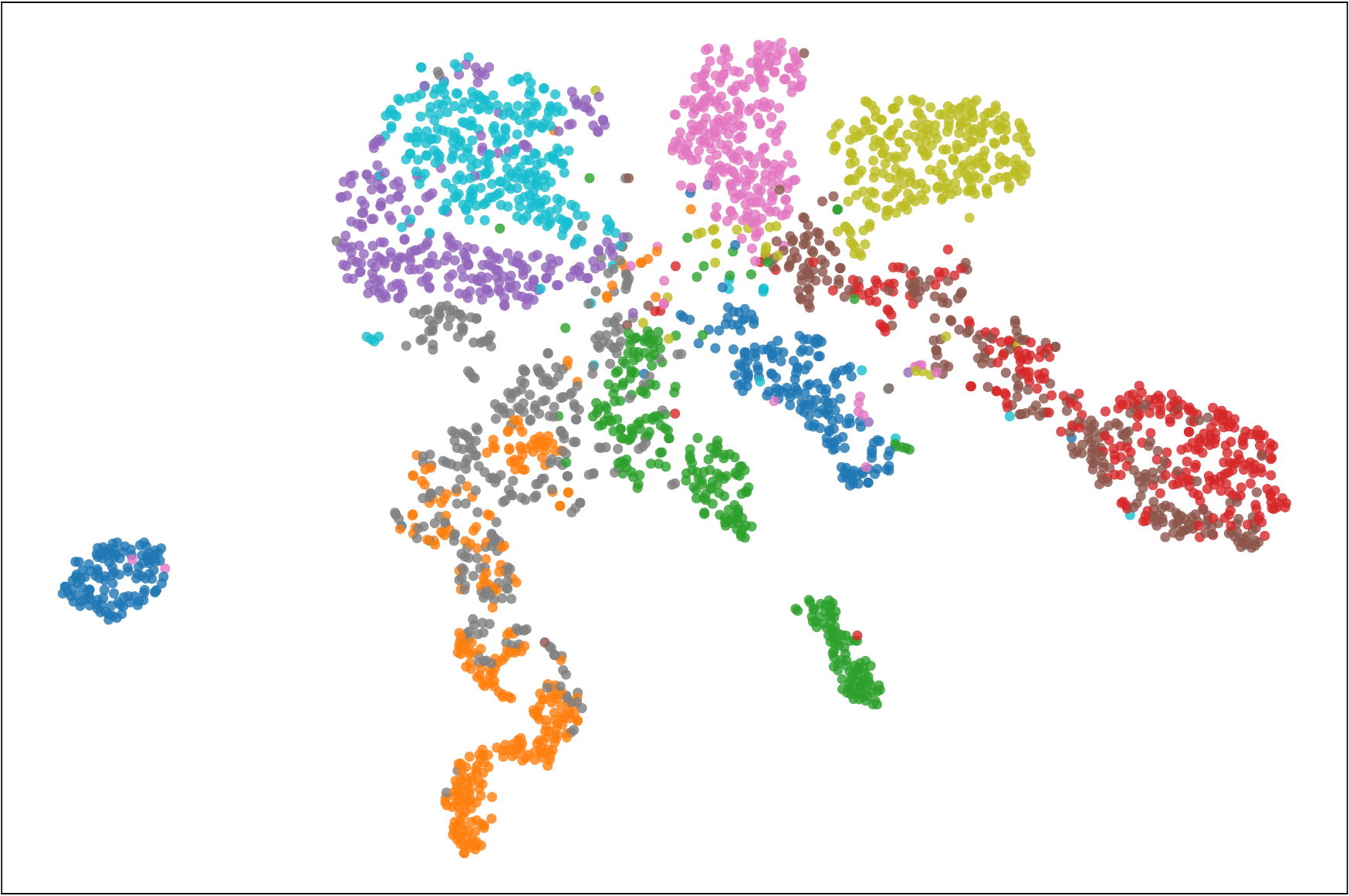} 
\\
{\footnotesize (a) SimCLR (Background)} & \hspace*{-0.2in} {\footnotesize (b) {\name} (Background)} & & \hspace*{-0.06in} {\footnotesize (c) SimCLR (Digit)} & \hspace*{-0.18in} {\footnotesize (d) {\name} (Digit)}
\end{tabular}
\vspace{-5pt}
\end{center}
\caption{\footnotesize{Visualization of latent features learned using different approaches on Colorful-Moving-MNIST dataset. The color of the left two figures corresponds to background object labels, and the color of the right two figures corresponds to the digit label.}}\label{fig:feat_vis}
\vspace{-10pt}
\end{figure*}

We report the main results for all datasets. The experiment setup, training details and hyper-parameter settings are provided in the supplemental material along with additional results. 

\textbf{ImageNet.} Table \ref{tab:imgnet} compares \name\ with the contrastive learning baselines on the task of object classification under different data augmentations. Here, we compare \name\ with SimCLR and MoCo since they use the same set of data augmentations. The results show that with fewer data augmentations, the accuracy of the contrastive learning baselines drops quickly due to \red{feature suppression}. For example, removing the color distortion augmentation significantly degrades the performance of the baseline approaches, as color distribution is known to be \red{able to suppress other features in} contrastive learning. In contrast, \name\ is significantly more robust. For example, with only random cropping, \name's Top-1 accuracy drops by only 6.9 whereas the Top-1 accuracy of SimCLR drops by 27.6 and the Top-1 accuracy of  MoCo drops by 12.1. \blue{We also compare \name\ with a \red{predictive} baseline \cite{pathak2016context}}. \blue{For the \red{predictive} baseline, though the model is not sensitive to different augmentations, the best performance is not comparable to contrastive learning, indicating \red{predictive} learning alone is not enough to learn fine-grained representations from images.}


\textbf{MPII.} We use \name\ and the contrastive learning baselines to learn representations from MPII, and evaluate them on the task of pose estimation. Table \ref{tab:result-pose} shows that \name\ improves the average PCKh (the standard metric for pose estimation) over the strongest contrastive baseline by 3.7 and achieves even higher gains on important keypoints such as Head and Wrist. This is because contrastive learning is likely to focus on \red{features} irrelevant to the downstream task, such as clothes and appearances.

\textbf{FairFace.}  Table \ref{tab:result-face} compares the contrastive learning baselines to \name~on the task of face-attribute classification.  The results show how contrastive learning struggles with multi-attribute classification. Specifically, the performance of the contrastive learning baselines on ethnicity classification is close to supervised learning of that attribute (62\% vs. 69\%). However, their results on age and gender classifications are significantly worse than supervised learning of those attributes (44\% and 78\% vs. 54\% and 91\%). This indicates that ethnicity \red{suppresses other features in} contrastive learning. This feature is partial since there are dependencies in how ethnicity manifests itself across age and gender. In contrast, \name\ is much more robust to such \red{feature suppression problem}, and its performance results on age and gender classifications are much closer to those of fully-supervised classification of those attributes.


\textbf{Colorful-Moving-MNIST.}  We use this dataset to further investigate how contrastive learning performs on multi-attribute classification. Recall that each image in this dataset contains a digit from MNIST on a randomly selected background object from the STL-10. We investigate whether the learned representation supports both digit and background classifications.  Table \ref{tab:result-toy} shows that the contrastive learning baselines learn only the task of background classification, and fail to learn a representation relevant to digit classification.  This shows that information related to the background prevents contrastive learning from capturing digit-relevant features. Note that the performance gap on digit classification between contrastive learning and supervised learning is very large (the accuracy is 15\% vs. 93\%). This is much larger than the gap we saw on FairFace because the information related to digit and background are totally independent, whereas features related to ethnicity, age, and gender have a significant overlap. 
In contrast, the representation learned by \name\ achieves very good accuracy on both background and digit classifications.

Figure~\ref{fig:feat_vis} provides a t-SNE visualization \cite{van2008visualizing} of the learned features for SimCLR and \name. The figure shows how \red{predictive} learning complement contrastive learning. Comparing Figures~\ref{fig:feat_vis}(c) and~\ref{fig:feat_vis}(d) reveals that \name's \red{predictive} branch allows it to capture information about digits that is lost in SimCLR. 

Finally, we run SimCLR and \name\ on Colorful-Moving-MNIST with different feature dimensions of 512 and 1024, as shown in Table \ref{tab:result-dim}. These results show that the performance of SimCLR does not change with larger dimensions. In fact, the same result can be seen from our theoretical analysis, which proves that when increasing the feature dimensions, contrastive learning experiences many local minimums that correspond to all of the global minimums of the lower dimensions, which tend to suppress features, while \name~can escape from those local minimums.
\begin{table}[h]
\centering
\vspace{4pt}
\caption{\footnotesize{Performance on Colorful-Moving-MNIST with different feature dimensions under different unsupervised methods.}}
\label{tab:result-dim}
\begin{threeparttable}
\resizebox{0.42\textwidth}{!}{
\begin{tabular}{c|c|c|c} 
\toprule[1.5pt]
\textsc{Method} & \begin{tabular}[c]{@{}c@{}}\textsc{Feature} \\ \textsc{Dimension}\end{tabular} & \begin{tabular}[c]{@{}c@{}}\textsc{Digit cls} \\ \textsc{Acc.} (\%) \end{tabular} & \begin{tabular}[c]{@{}c@{}}\textsc{Bkgd cls} \\ \textsc{Acc.} (\%) \end{tabular}  \\ 
\midrule\midrule
 \multirow{2}{*}{\begin{tabular}[c]{@{}c@{}}SimCLR \end{tabular}}&512                    & 16.0                                                                  & 48.4                                                                  \\
&1024                                                                                                         & 15.8                                                                 & 48.6                                                                  \\
\midrule
\multirow{2}{*}{\begin{tabular}[c]{@{}c@{}}\name \end{tabular}}&512                                                                                                          & 88.1                                                                  & 46.3                                                                  \\ 
&1024                                                                                                          & 88.2                                                                  & 46.5                                                                  \\ 

\bottomrule[1.5pt]
\end{tabular}
}
\end{threeparttable}
\end{table}

 \vspace{-5pt}
\section{Conclusion \& Limitations}
\vspace{-3pt}
In this paper, we introduce \red{predictive} contrastive learning (\name), a novel framework for making unsupervised contrastive learning more robust and allow it to preserve useful information in the presence of \red{feature suppression}. We theoretically analyze the reason why contrastive learning is vulnerable to \red{feature suppression}, and show that the \red{predictive} loss can help avoid \red{feature suppression} and preserve useful information. Extensive empirical results on a variety of datasets and tasks show that \name\ is effective at addressing the \red{feature suppression} problem. 

\red{The problem of feature suppression is complex; and, while \name\ provides an important improvement over the current SOTA, it has some limitations. First,  \name\ sees some performance drop with fewer augmentations. The drop however is much better than the contrastive baselines. Second, \name\ tries to abstract and preserve the information in the input, but some of this information may be unnecessary or irrelevant to the downstream tasks of interest. Yet, despite these limitations, we believe that \name~provides an important step forward toward making self-supervised learning more robust} and providing richer self-supervised representations that support multi-attribute classifications and generalize well across diverse tasks.

 {\small
 \bibliographystyle{ieee_fullname}
 \bibliography{main}
 }

\clearpage
\section*{Appendix A: Additional Results}
In this section, we provide additional results to better understand \name.

\begin{table}[h]
\centering
\caption{Comparison of different predictive tasks for PCL's predictive branch. The table shows 
the performance of \name~on Colorful-Moving-MNIST with different predictive tasks in its predictive branch for the fixed feature encoder setting. Colorization achieves good performance on background classification, but bad performance on digit classification since the MNIST digits have no RGB information. Inpainting achieves the best performance among these predictive tasks.}
\label{tab:ablation}
\begin{tabular}{l |@{\hspace{0.2cm}} c c @{\hspace{0.2cm}} c @{\hspace{0.2cm}}c  @{\hspace{0.2cm}}c}\toprule[1.5pt]

  & \multicolumn{2}{c}{Colorful-Moving-MNIST}  \\
\midrule
Recon. Tasks & \begin{tabular}[c]{@{}c@{}}\textsc{Digit cls} \\ \textsc{Acc.} (\%) \end{tabular} & \begin{tabular}[c]{@{}c@{}}\textsc{Bkgd cls} \\ \textsc{Acc.} (\%) \end{tabular}  \\
\midrule
No Recon. & 15.7 & \textbf{48.5}  \\
\midrule
Jigsaw Puzzle & 16.1 & 47.7 \\ 
Colorization & 63.9 & 47.0  \\
Autoencoder & 65.6 & 42.9  \\
Inpainting & \textbf{88.3} & 46.5 \\
\bottomrule[1.5pt]
\end{tabular}
\end{table}

\textbf{Comparison of Different Predictive Tasks for \name's Predictive Branch}: In \name, we choose the inpainting task for the predictive branch. However, other predictive tasks can be potentially used for the predictive branch. In this section, we evaluate the performance of \name~with different predictive tasks including Jigsaw Puzzle \cite{noroozi2016unsupervised}, inpainting, auto-encoder and colorization \cite{zhang2016colorful}. Table \ref{tab:ablation} shows \name's performance using different predictive task on Colorful-Moving-MNIST under the fixed feature extractor setting. As shown in the table, all predictive tasks except for Jigsaw Puzzle significantly reduce errors on digit classification in comparison to using contrastive learning without any predictive task. This is because the Jigsaw Puzzle task does not require the features to be able to reconstruct the original image, but just to restore the order of different patches. Learning the background object is sufficient to solve Jigsaw, and the network does not have incentives to learn the digit. The table also shows that inpaintaing compares favorably to other predictive tasks and achieves good performance on both downstream tasks. Hence, we use inpainting as the default predictive task in \name.

\begin{table}[h]
\centering
\caption{Performance of MoCo on Colorful-Moving-MNIST without masking augmentation (fixed feature encoder setting). The results demonstrate that simply adding masking as a data augmentation does not achieve similar improvements as \name.}
\label{tab:ablation-cutout}
\resizebox{0.48\textwidth}{!}{
\begin{tabular}{l |@{\hspace{0.2cm}} c  c  @{\hspace{0.2cm}}  c @{\hspace{0.2cm}}c  @{\hspace{0.2cm}}c}\toprule[1.5pt]
  & \multicolumn{2}{c}{Colorful-Moving-MNIST} \\
\midrule
Recon. Tasks & \begin{tabular}[c]{@{}c@{}}\textsc{Digit cls} \\ \textsc{Acc.} (\%) \end{tabular} & \begin{tabular}[c]{@{}c@{}}\textsc{Bkgd cls} \\ \textsc{Acc.} (\%) \end{tabular}  & \\

\midrule
MoCo w/o masking & 15.7 & \textbf{48.5} \\
MoCo w/ masking & 15.2 & 48.4 \\
\name & \textbf{88.3} & 46.5  \\
\bottomrule[1.5pt]
\end{tabular}
}
\end{table}

\textbf{Masking as a Data Augmentation vs. \name}: In the predictive branch, \name~introduces masked input images. Some may wonder whether the improvements are coming from this masking operation, since cutting out the input signals can be viewed as one way of augmentation \cite{chen2020simple}. However, here in Table \ref{tab:ablation-cutout}, we show the performance of MoCo with and without masking augmentation on Colorful-Moving-MNIST (we use the same masking strategy as the predictive branch of \name). As shown in the table, the performance of MoCo stays similar with or without masking augmentation. This demonstrate that the improvements of \name~do not come from this augmentation.

\begin{table*}[t]
\centering
\caption{Digit classification and background classification accuracy of \name~ with different $\lambda$ on Colorful-Moving-MNIST dataset.}
\vspace{-10pt}
\label{tab:ablation-lambda}
\resizebox{0.75\textwidth}{!}{
\begin{tabular}{c|c|c|c|c|c|c|c|c|c|c}
\toprule[1.5pt]
$\lambda$ & 0    & 1    & 5    & 10   & 25   & 50   & 100    & 200   & 500   & 1000 \\
\midrule
\textsc{Digit Acc (\%)}                  & 15.7 & 48.6 & 69.8 & 88.3 & 88.2 & 88.3 & 88.1 & 87.5 & 86.3 & 85.0 \\\midrule
\textsc{Bkgd Acc (\%)}                   & 48.5 & 47.9 & 47.5 & 47.2 & 47.2 & 47.1 & 47.0  & 45.7 & 44.5 & 40.5 \\
\bottomrule[1.5pt]
\end{tabular}
}
\end{table*}

\begin{table*}[t]
\centering
\vspace{-2pt}
\caption{\footnotesize{Performance of {\name} and predictive baselines on MPII for the downstream task of human pose estimation. $\uparrow$ indicates the larger the value, the better the performance.}}
\vspace{2pt}
\label{tab:result-pose}
\begin{threeparttable}
\resizebox{0.98\textwidth}{!}{
\begin{tabular}{c|c|c|c|c|c|c|c|c|c} 
\toprule[1.5pt]
\multicolumn{2}{c|}{\textsc{Metric}} & Head$^\uparrow$ & Shoulder$^\uparrow$ & Elbow$^\uparrow$ & Wrist$^\uparrow$ & Hip$^\uparrow$ & Knee$^\uparrow$ & Ankle$^\uparrow$ & PCKh$^\uparrow$ \\ 
\midrule
\midrule
\multirow{5}{*}{\begin{tabular}[c]{@{}c@{}}\textsc{Fixed}\\\textsc{feature}\\\textsc{extractor} \end{tabular}} 


&Inpainting  & 83.4 & 75.2 & 53.6 & 44.4 & 56.4 & 44.3 & 45.7 & 59.0  \\ 
&Colorization  & 79.5 & 71.2 & 49.6 & 42.1 & 54.2 & 40.7 & 41.9 & 55.1  \\ 
&Autoencoder  & 79.1 & 70.1 & 47.2 & 41.6 & 51.9 & 39.1 & 40.3 & 53.8  \\ 

\cmidrule{2-10}
&\bf \name~(ours)                                                                                      & \bf 85.7 & \bf 78.8 & \bf 61.7 & \bf 51.3 & \bf 64.4 & \bf 55.6 & \bf 49.2 & \bf 65.1                                                                                            \\ 
&\bf \textsc{Improvements}                                                                                      
& \bf \textcolor{darkgreen}{\textbf{+2.3}}
& \bf \textcolor{darkgreen}{\textbf{+3.6}} 
& \bf \textcolor{darkgreen}{\textbf{+8.1}} 
& \bf \textcolor{darkgreen}{\textbf{+6.9}} 
& \bf \textcolor{darkgreen}{\textbf{+8.0}} 
& \bf \textcolor{darkgreen}{\textbf{+11.3}} 
& \bf \textcolor{darkgreen}{\textbf{+3.5}}
& \bf \textcolor{darkgreen}{\textbf{+6.1}}                                                                                          \\ 
\midrule
\midrule
\multirow{5}{*}{\begin{tabular}[c]{@{}c@{}}\textsc{Fine-}\\\textsc{tuning} \end{tabular}} 
&Inpainting & 96.3 & \bf 95.2 & 87.9 & 82.1 & 87.8 & 82.5 & 77.6 & 87.7 \\ 
&Colorization & 96.2 & 95.1 & 87.7 & 82.1 & 87.8 & 82.5 & 77.5 & 87.6 \\ 
&Autoencoder & 96.0 & 94.9 & 87.6 & 82.0 & 87.6 & 82.4 & 77.3 & 87.5 \\ 

\cmidrule{2-10}
&\bf \name~(ours)                                                                                     & \bf 96.3 &94.9 & \bf 88.1 & \bf 82.3 & \bf 87.9 & \bf 82.8 & \bf 77.8 & \bf 87.8                                                                                                                                                                        \\
&\bf \textsc{Improvements}                                                                                      
& \bf \textcolor{darkgreen}{\textbf{+0.0}} 
& \bf \textcolor{lightblue}{\textbf{-0.3}}  
& \bf \textcolor{darkgreen}{\textbf{+0.2}} 
& \bf \textcolor{darkgreen}{\textbf{+0.2}}
& \bf \textcolor{darkgreen}{\textbf{+0.1}} 
& \bf \textcolor{darkgreen}{\textbf{+0.3}} 
& \bf \textcolor{darkgreen}{\textbf{+0.2}} 
& \bf \textcolor{darkgreen}{\textbf{+0.1}}                                                                                            \\ 


\bottomrule[1.5pt]
\end{tabular}}
\end{threeparttable}
%
%
\begin{tabular}{cc}

\begin{minipage}{.52\linewidth}
\centering
\caption{\footnotesize{Performance on FairFace with {\name} and  different predictive unsupervised learning methods. The models are evaluated on downstream tasks of age, gender and ethnicity classification.}}
\label{tab:result-face}
\begin{threeparttable}
\resizebox{.98\textwidth}{!}{
\begin{tabular}{c|c|c|c|c} 
\toprule[1.5pt]
\multicolumn{2}{c|}{\textsc{Metric}}                                                                                  & \begin{tabular}[c]{@{}c@{}}\textsc{Age cls} \\ \textsc{Acc.} (\%) \end{tabular} & \begin{tabular}[c]{@{}c@{}}\textsc{Gender cls} \\ \textsc{Acc.} (\%) \end{tabular} & \begin{tabular}[c]{@{}c@{}}\textsc{Ethn. cls} \\ \textsc{Acc.} (\%) \end{tabular} \\ 
\midrule\midrule
\multirow{6}{*}{\begin{tabular}[c]{@{}c@{}}\textsc{Fixed}\\\textsc{feature}\\\textsc{extractor} \end{tabular}} 
&Inpainting & 46.3  & 83.6  & 52.9  \\
&Colorization & 46.1  & 82.9  & 53.8  \\
&Autoencoder & 44.3  & 80.1  & 50.7  \\

\cmidrule{2-5}
&\bf{\name~(ours)}  & \bf{50.0}  & \bf{87.2}  & \bf 61.2     \\ 
&\bf \textsc{Improvement}                                                                                      & \bf \textcolor{darkgreen}{\textbf{+3.7}}  & \textcolor{darkgreen}{\textbf{+3.6}}  & \textcolor{darkgreen}{\textbf{+7.4}} \\ 
\midrule\midrule
\multirow{6}{*}{\begin{tabular}[c]{@{}c@{}}\textsc{Fine-}\\\textsc{tuning} \end{tabular}}
&Inpainting & 55.0  & 91.8  & 68.3  \\
&Colorization & 54.9  & 92.0  & 68.6  \\
&Autoencoder & 54.5  & 91.3  & 67.9  \\
\cmidrule{2-5}
&\bf{\name~(ours)}                                                                                    &  \textbf{55.3}                                                                  &  \textbf{92.3}                 & \bf 69.0                                                 \\ 
&\bf \textsc{Improvement}                                                                                     & \bf \textcolor{darkgreen}{\textbf{+0.3}}      & \bf \textcolor{darkgreen}{\textbf{+0.3}}                                                         & \textcolor{darkgreen}{\textbf{+0.4}}                                                             \\ 
\bottomrule[1.5pt]
\end{tabular}
}
\end{threeparttable}
\end{minipage}
&
\begin{minipage}{.44\linewidth}
\centering
\caption{\footnotesize{Performance on Colorful-Moving-MNIST under different methods. The models are evaluated on the downstream tasks of digit classification and background object classification.}}
\label{tab:result-toy}
\begin{threeparttable}
\resizebox{0.98\textwidth}{!}{
\begin{tabular}{c|c|c|c} 
\toprule[1.5pt]
\multicolumn{2}{c|}{\textsc{Metric}}                                                                                  & \begin{tabular}[c]{@{}c@{}}\textsc{Digit cls} \\ \textsc{Acc.} (\%) \end{tabular} & \begin{tabular}[c]{@{}c@{}}\textsc{Bkgd cls} \\ \textsc{Acc.} (\%) \end{tabular}  \\ 
\midrule\midrule
\multirow{6}{*}{\begin{tabular}[c]{@{}c@{}}\textsc{Fixed}\\\textsc{feature}\\\textsc{extractor} \end{tabular}} 
&Inpainting    & 84.7                  & 35.0\\     
&Colorization    & 80.7                  & 38.4\\      
&Autoencoder    & 81.0                  & 32.9\\      
 
\cmidrule{2-4}
&\bf{\name~(ours)}                                                                                      & \bf 88.3                                                                  & \bf 46.5                                                                  \\ 
&\bf \textsc{Improvement}                                                                                      & \bf \textcolor{darkgreen}{\textbf{+3.2}}                                                                & \textcolor{darkgreen}{\textbf{+8.1}} \\ 
\midrule\midrule
\multirow{6}{*}{\begin{tabular}[c]{@{}c@{}}\textsc{Fine-}\\\textsc{tuning} \end{tabular}} 
&Inpainting                                                                                                          & 92.9                                                                   & 54.5                                                                  \\ 
&Colorization & 92.5 & 54.5 \\
&Autoencoder & 92.4 & 54.1 \\

\cmidrule{2-4}
&\bf{\name~(ours)}                                                                                    & \bf 93.3                                                                   & \bf 54.7                                                                  \\ 
&\bf \textsc{Improvement}                                                                                     & \bf \textcolor{darkgreen}{\textbf{+0.4}}                                                               & \textcolor{darkgreen}{\textbf{+0.2}}                                                             \\ 
\bottomrule[1.5pt]
\end{tabular}
}
\end{threeparttable}


\end{minipage}

\end{tabular}

\end{table*}

\begin{table}[h]
\centering
\caption{Performance of \name~on Colorful-Moving-MNIST with and without warm-up training.}
\label{tab:result-warmup-rgb}
\begin{tabular}{l |@{\hspace{0.2cm}} c c  @{\hspace{0.2cm}}  c @{\hspace{0.2cm}}c  @{\hspace{0.2cm}}c}\toprule[1.5pt]
  & \multicolumn{2}{c}{Colorful-Moving-MNIST} \\
\midrule
Warm-up Training  & \begin{tabular}[c]{@{}c@{}}\textsc{Digit cls} \\ \textsc{Acc.} (\%) \end{tabular} & \begin{tabular}[c]{@{}c@{}}\textsc{Bkgd cls} \\ \textsc{Acc.} (\%) \end{tabular}   \\
\midrule
No & 24.9 & \textbf{47.8}  \\
Yes & \textbf{88.3} & 46.5\\
\bottomrule[1.5pt]
\end{tabular}
\vspace{-10pt}
\end{table}

\textbf{Warm-up Training}: To show the effectiveness of the proposed warm-up training strategy (Sec. 3 (c)), we compare the results of warm-up training with the results of directly using the combined loss $\mathcal{L}$ (i.e., combining the prediction loss and the contrastive loss) from the beginning. As shown in Table \ref{tab:result-warmup-rgb}, without the warm-up training, on Colorful-Moving-MNIST, \name~largely degenerates to become similar to the contrastive learning baselines and cannot learn good features related to digit classification. This indicates that without the warm-up phase, the contrastive loss can dominate the network causing it to suppress feature at the beginning, and that the network cannot later jump out of the local minimum that suppress feature. On the other hand, with warm-up training, the network first learns a coarse representation; then the contrastive loss helps the network learn more fine-grained representations.

\textbf{Performance of \name~with different $\lambda$}: In \name, the combined loss is a weighted average of the prediction loss and the contrastive loss, i.e., $\mathcal{L} = \mathcal{L}_{c} + \lambda  \cdot \mathcal{L}_{p}$. In the experiments of main paper, $\lambda$ is set to 10. In this section, we investigate how different $\lambda$ affects the  performance of \name. Note that when $\lambda=0$, \name~degenerates to contrastive learning; when $\lambda\rightarrow \infty$, \name~degenerates to predictive learning.

Table \ref{tab:ablation-lambda} compares the performance of \name~ with different $\lambda$. As we can see from the results, when $\lambda < 100$, with larger $\lambda$, the accuracy of the digit classification increases, while the accuracy of background classification decreases. Moreover, the $\lambda$ values between 10 and 100 gives quite similar performances, indicating a balancing between contrastive loss and prediction loss. For $\lambda>100$, the prediction loss dominates the contrastive loss and harm the performance. Therefore, we fix $\lambda=10$ for all experiments. 



\textbf{Predictive Learning vs. \name}: In the main paper, we mainly compare \name~with contrastive learning since contrastive learning is the current unsupervised learning SOTA on ImageNet and outperforms predictive learning by a large margin \cite{chen2020simple,grill2020bootstrap}. Here, we also compare \name~with predictive learning, such as inpainting, colorization and autoencoder, on various datasets to demonstrate the effectiveness of the contrastive branch of \name. 
Tables [4-6]
 compare {\name} with Inpainting \cite{pathak2016context}, Colorization \cite{zhang2016colorful} and Auto-encoder on the RGB datasets. The results demonstrate that \name~outperforms all predictive learning baselines by a large margin. This is because the contrastive branch in \name~can significantly improve the quality of the learned representation so it can achieve much better performance on downstream tasks.

\section*{Appendix B: Additional Proofs}
Here we formally prove Corollary 1 and 2.
\begin{corollary}\label{cor:stationary}
	For any lifting operator $\lift$, if $\hat{Z} = \{\hat{z_i}\}$ is a stationary point of $\ENCE(Z; X, \tau, d_1)$, then $\lift(\hat{Z})$ is a stationary point of $\ENCE(Z; X, \tau, d_2)$.
\end{corollary}
\begin{proof}
The proof is in the supplemental material. 
\end{proof}

\begin{proof}
	$\hat{Z}$ is a stationary point of $\ENCE(Z; X, \tau, d_1)$ implies $\nabla_{z_i} \ENCE(\hat{Z}; X, \tau, d_1)=0$. Therefore, by Lemma 2,
	$\nabla_{\tilde{z_i}}\ENCE(\lift(\hat{Z}); X, \tau, d_2) = \lift \left( \nabla_{z_i} \ENCE(\hat{Z}; X, \tau, d_1) \right) = 0$. 
\end{proof}

\begin{corollary}\label{cor:saddle}
	For any lifting operator $\lift$, if $\hat{Z} = \{\hat{z_i}\}$ is a global minimum of $\ENCE(Z; X, \tau, d_1)$ with a positive definite Hessian matrix, then $\lift(\hat{Z})$ is a saddle point or a local minimum of $\ENCE(Z; X, \tau, d_2)$.
\end{corollary}

\begin{proof}
	From Corollary \ref{cor:stationary}, $\lift(\hat{Z})$ is a stationary point of $\ENCE(Z; X, \tau, d_2)$. Since the Hessian matrix of $\ENCE(Z; X, \tau, d_1)$ at $\hat{Z}$ is positive definite, $\forall r>0, \exists Z'\in B_r(\hat{Z})$ s.t. $\ENCE(Z'; X, \tau, d_1) > \ENCE(Z; X, \tau, d_2)$, where $B_r(Z)=\{Z'\in\mathbb{S}^{d-1} | \text{ }||Z-Z'||_2<r \}$ is the neighborhood of $Z$ with radius $r$. Therefore, $\ENCE(\lift(Z'); X, \tau, d_1) > \ENCE(\lift(Z); X, \tau, d_2)$ (Lemma 1). Note that $Z'\in B_r(\hat{Z}) \rightarrow \lift(Z') \in B_r(\lift(\hat{Z}))$. Therefore, $\forall r>0, \exists \lift(Z')\in B_r(\lift(\hat{Z}))$ s.t. $\ENCE(\lift(Z'); X, \tau, d_1) > \ENCE(\lift(Z); X, \tau, d_2)$. Therefore, $\lift(\hat{Z})$ is not a local maximum, so it can only be a local minimum or a saddle point of $\ENCE(Z; X, \tau, d_2)$. 
\end{proof}
\section*{Appendix C: Implementation Details}
In this section, we provide the implementation details of the models used in our experiments. All experiments are performed on 8 NVIDIA Titan X Pascal GPUs. On ImageNet, training takes $\sim$100 hours. On each dataset, we fix the batch size and training epochs for different baselines for a fair comparison. Other parameters for each baseline follow the original paper to optimize for its best performance. Code will also be released upon acceptance of the paper.

\textbf{ImageNet:} We use a standard ResNet-50 for the encoder network. The decoder network is a 11-layer deconvolutional network. The projection head for contrastive learning is a 2-layer non-linear head which embeds the feature into a 128-dimensional unit sphere. The same network structure is used for all baselines and \name. 

We follow the open repo of \cite{chen2020improved} for the implementation of MoCo baselines and \name~on ImageNet. For results on SimCLR, we follow the results reported in \cite{grill2020bootstrap}. All baselines and \name~is trained for 800 epochs with a batch size of 256. For the predictive branch of \name~and the predictive baseline, we mask out $3$ to $5$ rectangles at random locations in the image. The size of each square is chosen by setting its side randomly between 40 and 80 pixels. For the contrastive branch of \name, we apply the same training scheme as MoCo. The first 10 epochs are warm-up epochs, where we only train the network with the prediction loss $\mathcal{L}_p$. For later training, we set $\lambda=10$. For other RGB datasets, we mainly follows similar implementation as ImageNet.

\textbf{MPII:} We use the network structure similar to the one in \cite{xiao2018simple}. We use a ResNet-50 for the encoder network. Three deconvolutional layers with kernel size 4 and one convolutional layer with kernel size 1 is added on top of the encoded feature to transfer the feature into 13 heatmaps corresponding to 13 keypoints. For the contrastive branch, a 2-layer non-linear projection head is added on top of the encoded feature and embeds the feature into a 128-dimensional unit sphere. For the predictive branch, a decoder network similar to the pose estimation deconvolution network (only the number of output channels is changed to 3) is used to reconstruct the original image. Other implementation details are the same as ImageNet. 

For the baselines and \name, we train the network for 300 epochs with a batch size of  256. The data augmentation is the same as the baseline augmentations on ImageNet. 
For \name, the first 10 epochs are warm-up epochs, where we only train the network with the prediction loss $\mathcal{L}_p$.

\textbf{FairFace:} We use a standard ResNet-50 for the encoder network. The decoder network is a 11-layer deconvolutional network. The projection head for contrastive learning is a 2-layer non-linear head which embeds the feature into a 128-dimensional unit sphere. The same network structure is used for all baselines and \name. 

For the baselines and \name, we train the network for 1000 epochs with a batch size of  256. The data augmentation is the same as the baseline augmentations on ImageNet. 
For \name, the first 30 epochs are warm-up epochs, where we only train the network with the prediction loss $\mathcal{L}_p$.

\textbf{Colorful-Moving-MNIST:} We use a 6-layer ConvNet for the encoder. The encoder weights for the predictive and contrastive branches are shared. The decoder is a 6-layer deconvolutional network symmetric to the encoder. The projection head for contrastive learning is a 2-layer non-linear head which embeds the feature into a 64-dim normalized space. 

We use the SGD optimizer with 0.1 learning rate, 1e-4 weight decay, and 0.9 momentum to train the model for 200 epochs. The learning rate is scaled with a factor of 0.1 at epoch 150 and 175. The batch size is set to 512. The temperature for contrastive loss is set to 0.1. For \name, the first 30 epochs are warm-up epochs, where we only train the network with the prediction loss $\mathcal{L}_p$.

For \name, for each input image after augmentation with a size of $64$ by $64$ pixels, we randomly mask out $3$ to $5$ rectangle patches at random locations in the image and fill them with the average pixel value of the dataset. The size of each square is chosen by setting its side randomly between $10$ and $16$ pixels.

\section*{Appendix D: Experiments' Setup and Evaluation Metrics}

\noindent\textbf{Setup.} On ImageNet, as common in the literature, we evaluate the representations with the encoder fixed and only the linear classifier is trained. On all other datasets, we evaluate the representations under two different settings: fixed feature encoder setting and fine-tuning setting. In the fixed feature encoder setting, the ResNet encoder is fixed and only the classifier (FairFace, Colorful-Moving-MNIST) or the 4-layer decoder network (MPII); In the fine-tuning setting, the encoder is initialized with the pre-trained model and fine-tuned during training.

\noindent\textbf{Evaluation Metrics.}
For ImageNet, FairFace and Colorful-Moving-MNIST, the evaluation metrics are the standard Top-1 classification accuracy. For MPII, we evaluate the learned representations under the single pose estimation setting~\cite{andriluka20142d}. Each person is cropped using the approximate location and scale provided by the dataset. Similar to prior works, we report the PCKh (Percentage of Correct Keypoints that uses the matching threshold as 50\% of the head segment length) value of each keypoint and an overall weighted averaged PCKh over all keypoints (head, shoulder, elbow, wrist, hip, knee, ankle).


\end{document}